\documentclass[12pt]{article} 

\usepackage{amsmath, amsfonts, amssymb}
\usepackage[margin=0.75in]{geometry}
\usepackage{graphicx}
\usepackage{hyperref}
\usepackage{setspace}
\usepackage{algorithm}
\usepackage{algpseudocode}
\usepackage{tabularx}
\usepackage{lscape} 
\usepackage{rotating}
\usepackage{fancyhdr}
\usepackage{hyphenat} 
\usepackage{enumitem}
\usepackage{mathrsfs}
\usepackage{bm}
\usepackage{placeins} 
\usepackage{cleveref}
\usepackage{fancyhdr}
\usepackage{color}
\usepackage{lineno}
\usepackage{multirow}
\usepackage{adjustbox} 
\usepackage{tikz} 
\usepackage{longtable} 
\usepackage{caption}
\usepackage{graphicx}
\usepackage{makecell}
\newcommand{\edit}[1]{\color{black}#1 \color{black}}

\title{ADAM-SINDy: An Efficient Optimization Framework for Parameterized Nonlinear Dynamical System Identification}
\author{Siva Viknesh$^{1,2}$ \and Younes Tatari$^{1,2}$ \and Chase Christenson$^{2}$ \and  Amirhossein Arzani$^{1,2}$ }
\date{}

\begin{document}
\maketitle

\begin{center}
$^1$Department of Mechanical Engineering, University of Utah, Salt Lake City, UT, USA \\
$^2$Scientific Computing and Imaging Institute, University of Utah, Salt Lake City, UT, USA\\
\end{center}

\bigskip

\noindent Correspondence:\\
Amirhossein Arzani,\\
University of Utah,\\
Salt Lake City, UT,  84112\\
Email: amir.arzani@sci.utah.edu

\thispagestyle{empty}


\begin{abstract}


Identifying nonlinear dynamical systems characterized by nonlinear parameters presents significant challenges in deriving mathematical models that enhance understanding of physical phenomena. Traditional methods, such as Sparse Identification of Nonlinear Dynamics (SINDy) and symbolic regression, can extract governing equations from observational data; however, they also come with distinct advantages and disadvantages. This paper introduces a novel methodology within the SINDy framework, termed ADAM-SINDy, which synthesizes the strengths of established approaches by employing the ADAM optimization algorithm. This integration facilitates the simultaneous optimization of nonlinear parameters and coefficients associated with nonlinear candidate functions, enabling efficient and precise parameter estimation without requiring prior knowledge of nonlinear characteristics such as trigonometric frequencies, exponential bandwidths, or polynomial exponents, thereby addressing a key limitation of the classical SINDy framework. Through an integrated global optimization, ADAM-SINDy dynamically adjusts all unknown variables in response to system-specific data, resulting in a more adaptive and efficient identification procedure that reduces the sensitivity to the library of candidate functions. The performance of the ADAM-SINDy methodology is demonstrated across a spectrum of dynamical systems, including benchmark coupled nonlinear ordinary differential equations such as oscillators, chaotic fluid flows, reaction kinetics, pharmacokinetics, as well as nonlinear partial differential equations (wildfire transport). The results demonstrate significant improvements in identifying parameterized dynamical systems and underscore the importance of concurrently optimizing all parameters, particularly those characterized by nonlinear parameters. These findings highlight the potential of ADAM-SINDy to extend the applicability of the SINDy framework in addressing more complex challenges in dynamical system identification.


\end{abstract}

\newpage

\section{Introduction}
\label{sec1}

System identification is a crucial aspect of accurately modeling the dynamics of physical, chemical, and biological systems. By providing a mathematical representation of these systems, system identification enables essential predictions, interpretation, optimization, and control strategies across diverse applications, from engineering to environmental science~\cite{ljung2010perspectives}. The process typically involves determining a system's mathematical model from observed data, using methods such as least squares, subspace identification, or frequency domain analysis. The process includes data collection, model structure selection, parameter estimation, validation, and refinement.

Machine learning significantly enhances system identification by offering advanced tools for modeling complex, nonlinear systems and uncovering hidden patterns in data. Techniques such as supervised learning, reinforcement learning, neural networks, and Gaussian processes facilitate flexible, data-driven modeling. The integration of machine learning with traditional system identification has led to the development of methodologies for linear~\cite{van2012subspace}, nonlinear~\cite{nelles2020nonlinear}, hybrid~\cite{lauer2018hybrid}, and parametric/non-parametric~\cite{pai2008hht} dynamical systems. Although linear models are often simpler and computationally efficient, practical systems frequently contain nonlinear features that necessitate more sophisticated models. Addressing the challenge of defining appropriate models for nonlinear systems often requires expert insight. Despite significant progress in earlier research~\cite{kumpati1990identification, suykens2012artificial}, nonlinear system identification remains a vibrant and promising research area. The increasing availability of data and advancements in computational methods have led to the development of automated techniques for nonlinear system identification. Recent advancements include methods such as genetic algorithms~\cite{gondhalekar2009parameters}, symbolic regression~\cite{quade2016prediction}, and deep learning-based algorithms~\cite{both2021deepmod, okunev2021ordinary, okunev2024nonlinear, stephany2022pde, rajendra2020modeling}, which have proven effective in deriving governing nonlinear equations from empirical data. These methods aim to generate meaningful models from data with minimal reliance on prior assumptions, making them particularly valuable when traditional model hypotheses are unavailable.

An important aspect of data-driven system identification is the interpretability of the identified system, which is often accomplished through the use of a limited number of terms in an analytical equation. This characteristic not only establishes a meaningful connection between system identification studies and explainable artificial intelligence (XAI) but also underscores the significance of transparency in modeling complex phenomena. In the realm of XAI, interpretable models are developed to explain the decisions of opaque neural networks (post-hoc XAI) or to enhance interpretability from the construction phase of the machine learning architecture (by-design XAI)~\cite{samek2021explaining,thampi2022interpretable,zhong2022explainable}. Importantly, the system identification methods presented in this study exemplify the principles of by-design XAI.

Symbolic regression provides a unique and interpretable method for system identification, blending classical equation-based approaches with advanced optimization techniques to produce insightful models~\cite{koza1994genetic}. Currently, the use of genetic algorithms represents the state-of-the-art in symbolic regression~\cite{bongard2007automated, schmidt2009symbolic}. These algorithms effectively mimic natural selection by evolving a population of candidate models over successive generations. Through essential operations such as crossover, mutation, and selection, genetic algorithms efficiently search for equations that optimally capture the system dynamics. This comprehensive process not only facilitates the simultaneous discovery of both the structure and parameters of governing equations but also proves particularly effective in capturing intricate system behaviors. However, it is often susceptible to the bloating effect~\cite{dick2014bloat}, where candidate solutions become overly complex, hindering interpretability and reducing the generalizability of the identified models. Despite its strengths, symbolic regression faces challenges in capturing chaotic systems due to difficulties in effective phase space sampling, often mitigated through embedding methods~\cite{sauer1991embedology, takens2006detecting}. Advancements such as diffusion maps~\cite{tenenbaum2000global, singer2009detecting} and local linear embedding~\cite{roweis2000nonlinear}, address the curse of dimensionality by operating directly on the manifold of the dynamical system, improving the ability to capture complex behaviors. Moreover, innovations like linear basis initialization~\cite{mcconaghy2011ffx,  burlacu2020operon}, fast function extraction~\cite{quade2016prediction}, gradient-based optimization~\cite{kronberger2020identification}, invoking system symmetries/manifolds~\cite{schmidt2009distilling}, and physics-constrained optimization~\cite{oh2024inherently} have significantly enhanced the performance of symbolic regression in predicting and identifying dynamical systems.

As an alternative to symbolic regression, sparsity-promoting techniques that represent nonlinear equations as a sparse combination of basis functions from a comprehensive dictionary have been proposed, offering a computationally efficient alternative~\cite{wang2011predicting, schaeffer2013sparse}. The Sparse Identification of Nonlinear Dynamics (SINDy) framework~\cite{brunton2016discovering} is a widely recognized methodology for discovering interpretable and parsimonious mathematical models that balance accuracy and simplicity. It is based on the principle that many physical dynamical systems can be described by parsimonious models, where only a small set of terms is needed to represent the system's dynamics. By framing the identification problem as a sparse regression problem, SINDy facilitates the discovery of compact, interpretable models while avoiding the computational intensity associated with symbolic regression. This sparsity-driven approach mitigates the risk of overfitting and simplifies the model selection process.

Unlike the broad unstructured search employed in symbolic regression, SINDy operates within a constrained framework where candidate functions are drawn from a predefined library of candidate terms. This structured approach significantly reduces the computational burden. It allows for a more targeted and controlled exploration of the governing equations, ensuring that only the most relevant terms are retained in the final model. This methodology not only enhances computational efficiency but can also improve interpretability. Furthermore, SINDy has demonstrated its effectiveness across various domains, including fluid dynamics~\cite{loiseau2018constrained}, plasma dynamics~\cite{dam2017sparse}, turbulence closures~\cite{beetham2020formulating}, chaotic systems~\cite{kaptanoglu2023benchmarking}, and computational chemistry~\cite{boninsegna2018sparse}. Additionally, SINDy has been extended to applications such as nonlinear model predictive control~\cite{kaiser2018sparse}, rational functions~\cite{kaheman2020sindy, goyal2022discovery}, enforcing conservation laws and symmetries~\cite{loiseau2018constrained}, interpretable operator learning~\cite{arzani2024interpreting}, promoting stability~\cite{kaptanoglu2021promoting}, and generalizing for stochastic dynamics~\cite{callaham2021nonlinear}, including Bayesian perspectives~\cite{zhang2018robust}.


Comparing symbolic regression and the SINDy methodologies for the identification of dynamical systems reveals several key differences that impact their application and efficiency. Symbolic regression does not necessitate prior knowledge of parameters associated with the complex behavior of a system, such as Fourier frequencies and exponential growth/decay rates. Instead, it searches through a vast space of mathematical operators and nonlinear functions, such as polynomials, trigonometrics, and logarithms, to construct an equation that best fits the given data. However, this exploratory nature of symbolic regression often entails considerable computational demands, necessitating extensive computational resources. Moreover, it may yield varying results for identical datasets and hyperparameters, posing challenges for reproducibility.  In contrast, the SINDy framework \textit{does} require prior knowledge of nonlinear parameters such as frequencies to effectively guide the optimization process. Moreover, efforts to make these parameters trainable during optimization, such as the Newton-gradient-based approach attempted in~\cite{champion2020unified}, have faced significant challenges due to optimization stiffness and the \textit{non-convexity} of the loss landscape, ultimately limiting their success. Without such preliminary information, SINDy may struggle to yield an accurate mathematical model, potentially leading to sub-optimal performance. Nevertheless, when the necessary parameters are known or when a distribution of parameter ranges is considered that contains the correct parameter, SINDy offers a considerable computational advantage. It can derive the desired mathematical model in a fraction of the computational time required by symbolic regression. This efficiency is attributed to its formulation of the identification problem as a sparse regression task and a convex optimization problem, which streamlines the model selection process and reduces computational overhead. 

While both symbolic regression and SINDy offer distinct advantages and limitations, their comparative strengths highlight the importance of continued advancements in nonlinear system identification. In this work, we adopted an optimization algorithm leveraging Stochastic Gradient Descent--ADAM optimization, implemented using the PyTorch machine learning framework, to significantly improve the classical SINDy methodology. This augmented approach, termed ADAM-SINDy, integrates key advantages of symbolic regression, such as the simultaneous optimization of nonlinear parameters (e.g., frequencies and exponential growth/decay rates), alongside the selection of the most appropriate candidates from a comprehensive library of functions. Its effectiveness stems from leveraging the computational graph of the gradient descent algorithm for efficient backpropagation and dynamic interactions between all the trainable parameters, thus enhancing the robustness and precision of system identification, particularly in complex dynamical systems where the traditional SINDy method struggles. It is important to note that while ADAM-SINDy is not designed to surpass symbolic regression in all aspects, its primary goal is to address the limitations of the classical SINDy framework, particularly in nonlinear parameter estimation. By enhancing SINDy's capability to simultaneously identify nonlinear parameters and select candidate functions, ADAM-SINDy offers a more efficient and scalable approach to identifying parameterized dynamical systems. This improvement broadens the applicability of the SINDy framework, making it more effective for complex and high-dimensional systems.

Our present work makes several key contributions:
\begin{itemize}
    \item The SINDy framework is augmented with ADAM optimization, termed ADAM-SINDy, which enables the simultaneous estimation of nonlinear parameters while identifying parsimonious and relevant candidate terms. This enhancement addresses a significant limitation of the classical SINDy method, which traditionally requires exact knowledge of the parameters in its library of candidate terms.  
    
    \item The ADAM-SINDy framework integrates concurrent hyperparameter optimization (the sparsity knob) during the identification process, thereby mitigating the need for extensive manual tuning. In contrast, the classical SINDy approach, although capable of producing parsimonious models, is sensitive to the sparsity knob value, necessitating a delicate balance between accuracy and model complexity.
    
    \item ADAM-SINDy introduces a candidate-wise sparsity knob,  adopting a strategy inspired by the Iteratively Reweighted Least Squares (IRLS) methodology~\cite{candes2008enhancing, daubechies2010iteratively, lai2013improved}. This approach selectively penalizes incorrect terms while retaining the relevant ones throughout the optimization process, employing a concurrent, adaptive individual weighting strategy that is absent in the classical SINDy method.

    \item Through the integration of a global optimization strategy, the ADAM-SINDy framework allows self-adaptation of nonlinear functions and interaction between candidate terms. This adaptability is crucial when predefined candidates are lacking, as it optimizes parameters to compensate for missing functions. In contrast, the classical SINDy approach is limited by static parameters and the absence of interaction, increasing the risk of incorrect system models when essential candidate functions are missing.
    
    \item The novel ADAM-SINDy optimization methodology allows for the incorporation of differential equation-based loss functions, akin to high-order Tikhonov regularization. This methodology provides an environment that can directly incorporate governing differential equations in certain types of problems into the optimization process, thereby accelerating loss decay and improving convergence rates.

\end{itemize}

The paper is structured as follows. Section~\ref{sec2} provides an overview of the classical SINDy methodology, outlining its mathematical foundation and then introducing the ADAM-SINDy methodology, emphasizing its advantages and the implementation of the gradient descent algorithm. This section also details the master candidate library of nonlinear functions used in evaluating the classical SINDy and ADAM-SINDy frameworks. Section~\ref{sec3} examines benchmark problems across various applications, including ordinary and partial differential equations, focusing on problems that have nonlinear parameters, thereby challenging the classical SINDy method.  Section~\ref{sec4} discusses the advantages of the ADAM-SINDy method over the classical approach. Section~\ref{sec5} presents a conclusion. \edit{In the Appendix, we provide more details for the algorithm and provide theoretical and empirical justification for the specific optimization formulation used in ADAM-SINDy. }

\section{Methods}\label{sec2}

\subsection{Sparse Identification of Nonlinear Dynamics -- SINDy }
The SINDy algorithm is an efficient nonlinear system identification approach, grounded in the principle that the governing equations of a nonlinear system can be effectively represented by a sparse combination of appropriate pre-defined candidate basis functions~\cite{brunton2016discovering}. The central objective of SINDy is to identify a minimal set of relevant basis functions from an extensive dictionary of candidate functions, leveraging sparsity--promoting techniques to achieve an optimal balance between model complexity and accuracy.

Consider the problem of identifying nonlinear dynamical systems described by the differential equation
\begin{equation}
\dot{\mathbf{x}}(t) = \mathbf{f}(\mathbf{x}(t)) \;,
\label{eq:system_dynamics}
\end{equation}
where \(\mathbf{x}(t) \in \mathbb{R}^{n}\) denotes the state vector at time \(t\), \(\dot{\mathbf{x}}(t)\) represents its time derivative, and $\mathbf{f}(\mathbf{x})$$:$ $\mathbb{R}^{n} \rightarrow \mathbb{R}^{n}$ is a nonlinear function governing the system's dynamics. The objective of the SINDy algorithm is to identify the function \(\mathbf{f}(\mathbf{x}(t))\) from available data. The process begins by collecting a dataset consisting of measurements of \(m\) states at \(n\) discrete time-steps. The data is organized as follows:

\begin{equation}
\mathbf{X} = 
\begin{bmatrix}
\mathbf{x}\left(t_{1}\right)^{\top} \\
\mathbf{x}\left(t_{2}\right)^{\top} \\
\vdots \\
\mathbf{x}\left(t_{n}\right)^{\top}
\end{bmatrix}
= 
\begin{bmatrix}
\mathbf{x}_{1}\left(t_{1}\right) & \mathbf{x}_{2}\left(t_{1}\right) & \cdots & \mathbf{x}_{m}\left(t_{1}\right) \\
\mathbf{x}_{1}\left(t_{2}\right) & \mathbf{x}_{2}\left(t_{2}\right) & \cdots & \mathbf{x}_{m}\left(t_{2}\right) \\
\vdots & \vdots & \ddots & \vdots \\
\mathbf{x}_{1}\left(t_{n}\right) & \mathbf{x}_{2}\left(t_{n}\right) & \cdots & \mathbf{x}_{m}\left(t_{n}\right)
\end{bmatrix} \;,
\end{equation}
where each row represents a snapshot of the states at a given time. The matrix \(\mathbf{X} \in \mathbb{R}^{n \times m}\) contains the state vectors \(\mathbf{x}(t_i)\), each corresponding to the system's state at time \(t_i\). Similarly, the time derivatives of the state vectors, denoted as \(\dot{\mathbf{X}} \in \mathbb{R}^{n \times m}\), are represented as follows:
\begin{equation}
\dot{\mathbf{X}} = 
\begin{bmatrix}
\dot{\mathbf{x}}\left(t_{1}\right)^{\top} \\
\dot{\mathbf{x}}\left(t_{2}\right)^{\top} \\
\vdots \\
\dot{\mathbf{x}}\left(t_{n}\right)^{\top}
\end{bmatrix}
= 
\begin{bmatrix}
\dot{\mathbf{x}}_{1}\left(t_{1}\right) & \dot{\mathbf{x}}_{2}\left(t_{1}\right) & \cdots & \dot{\mathbf{x}}_{m}\left(t_{1}\right) \\
\dot{\mathbf{x}}_{1}\left(t_{2}\right) & \dot{\mathbf{x}}_{2}\left(t_{2}\right) & \cdots & \dot{\mathbf{x}}_{m}\left(t_{2}\right) \\
\vdots & \vdots & \ddots & \vdots \\
\dot{\mathbf{x}}_{1}\left(t_{n}\right) & \dot{\mathbf{x}}_{2}\left(t_{n}\right) & \cdots & \dot{\mathbf{x}}_{m}\left(t_{n}\right)
\end{bmatrix} \;.
\end{equation}

This dataset is utilized as input to the system identification problem. A library of candidate functions, denoted \(\bm{\Theta}(\mathbf{X}, \bm{\Lambda}) \in \mathbb{R}^{n \times p}\), is constructed, encompassing $p$ number of functions, such as polynomials and trigonometric functions. A typical representative library is given below

\begin{equation}
\bm{\Theta}(\mathbf{X} \textbf{;}\thinspace \bm{\Lambda}) = \begin{bmatrix}
1  & \mathbf{X}^\mathcal{A} & \dots & \sin(\mathcal{B}\mathbf{X}) & \dots & \mathbf{X}\otimes \exp(\mathcal{C}\mathbf{X})
\end{bmatrix} \;,
\label{candidate}
\end{equation}
where each column of $\bm{\Theta}(\mathbf{X}\textbf{;}\thinspace \bm{\Lambda}) $ represents a candidate function derived from the state vector $\mathbf{X}$ for the given nonlinear parameters $\bm{\Lambda} \in \{\mathcal{A}, \mathcal{B}, \mathcal{C}\}$, which correspond to the highest polynomial order, Fourier frequencies, and exponential exponents, respectively, typically chosen to define the extensive candidate library. Here, the symbol, $\otimes$, indicates that column-wise Kronecker product of two candidate functions. These nonlinear parameters must be either predetermined prior to the optimization process, which is often impractical in real-world engineering problems, or defined within a specified distribution range, which can result in an extensive candidate matrix and risk missing the precise nonlinear parameter. Accurate determination of these parameters is a crucial step for constructing the candidate functions and effectively identifying the most relevant functions for characterizing the system's dynamics. 

The goal in SINDy is to select a subset of these candidate functions, which is achieved by formulating a sparse regression problem and aiming to determine a sparse corresponding coefficient vector \(\bm{\Xi} = \left[\bm{\xi}_{1}, \bm{\xi}_{2}, \ldots, \bm{\xi}_{m}\right] \in \mathbb{R}^{p\times m}\) indicating the active features from the dictionary $\bm{\Theta}(\mathbf{X}\textbf{;}\thinspace \bm{\Lambda})$, such that their weighted sum accurately describes \(\mathbf{f}(\mathbf{x}(t))\). Mathematically, this is expressed as

\begin{equation}
\dot{\mathbf{X}} = \bm{\Theta}(\mathbf{X}\textbf{;}\thinspace \bm{\Lambda}) \thinspace \bm{\Xi} \;.
\end{equation}

Sparsity-promoting regularization methods, such as LASSO~\cite{hastie2009elements, tibshirani1996regression}, are commonly used to enforce sparsity in \(\bm{\Xi}\), i.e.,\ making most of its entries zero. However, LASSO does not always guarantee the sparsest solution, and alternative methods~\cite{beck2013sparsity, yang2016sparse} may require prior knowledge of the expected number of non-zero elements in $\bm{\Xi}$, which is often unknown. The LASSO augmented with a sequential thresholding approach~\cite{brunton2016discovering}, which iteratively solves least-squares problems and prunes coefficients below a given threshold, offers a more flexible alternative. The sparse regression problem can be formulated as

\begin{equation}
\min_{\bm{\Xi}} \left[ \|\dot{\mathbf{X}} - \bm{\Theta}(\mathbf{X}\textbf{;}\thinspace \bm{\Lambda}) \thinspace \bm{\Xi}\|_2^2 + \lambda \|\bm{\Gamma}\thinspace\bm{\Xi}\|_1 \right] \;,
\label{optim_eq}
\end{equation}

where \(\lambda\) is a regularization parameter that enforces sparsity in the solution, balancing model complexity with accuracy. In the classical SINDy methodology~\cite{brunton2016discovering}, \(\bm{\Gamma} \in \mathbb{R}^{p\times p}\) is typically assumed to be an identity matrix. However, in more generalized settings where \(\bm{\Gamma}\) is a non-identity matrix, it serves to penalize individual candidate functions, either remaining fixed throughout the optimization process, as in the Adaptive LASSO~\cite{zou2006adaptive}, or being iteratively updated during optimization, as demonstrated in Iteratively Reweighted Least Squares (IRLS) method~\cite{candes2008enhancing, daubechies2010iteratively, lai2013improved}. Furthermore, if \(\bm{\Gamma}\) represents a differential equation, such as a gradient or Laplacian, the regularization approach resembles higher-order Tikhonov regularization~\cite{aster2018parameter}. Upon solving, the SINDy identified dynamical model is expressed as:

\begin{equation}
\dot{\mathbf{x}}(t) = \bm{\Theta}(\mathbf{x}(t)) \bm{\Xi} \;.
\end{equation}

The SINDy approach provides a parsimonious and interpretable framework for uncovering the underlying dynamics of a system by exploiting sparsity within a known function basis. This methodology is particularly advantageous when the governing equations are not fully known but are presumed to be sparse within the space of candidate functions.

\subsection{ADAM Optimization Augmented SINDy: ADAM-SINDy}
The ADAM-SINDy methodology, a novel extension of the traditional SINDy framework~\cite{brunton2016discovering}, introduces substantial improvements by adopting ADAM optimization in place of the more commonly employed iterative least squares algorithm for solving the LASSO problem. This methodological innovation significantly addresses key challenges in parameter estimation, while concurrently reducing the dependency on the predefined library of candidate functions, a key limitation that has often constrained the SINDy approaches. Unlike the classical SINDy framework, which heavily relies on the initial selection of candidate parameters, ADAM-SINDy \textit{adaptively} identifies relevant terms from the candidate library $(\bm{\Xi})$ while simultaneously optimizing nonlinear parameters $(\bm{\Lambda})$. This adaptive process minimizes reliance on prior assumptions and enhances the capability to discover complex, nonlinearly defined parameters in a more robust manner. Nonlinear parameters, such as frequencies in trigonometric functions, bandwidths in exponential functions (e.g., kernels), and non-integer polynomial exponents, present challenges in the traditional SINDy approach, which requires prior knowledge of these parameters. ADAM-SINDy addresses this limitation by making these parameters optimizable within a global optimization framework. \edit{
The sparse regression problem in Eq.~\ref{optim_eq} could be reformulated as a min-max or a joint-minimization optimization problem. The min-max optimization problem is written as
\begin{equation}
\min_{\bm{\Xi}, \bm{\Lambda}} \max_{\lambda, \bm{\Gamma}} \left[ \|\dot{\mathbf{X}} - \bm{\Theta}(\mathbf{X}; \bm{\Lambda}) \bm{\Xi}\|_2^2 + \lambda \| \;|\bm{\Gamma}| \; \bm{\Xi}\|_1 \right] \;,
\label{optim_eq_adam}
\end{equation}
where the objective is minimized with respect to the coefficient matrix \(\bm{\Xi}\) and nonlinear parameters \(\bm{\Lambda}\), and maximized with respect to sparsity-controlling variables \(\lambda\) and \(\bm{\Gamma}\). This min-max approach is similar to the adaptive weighting approach in physics-informed neural networks where the weights for different losses is adaptively updated~\cite{mcclenny2023self}. Alternatively, the problem can also be expressed as a joint minimization
\begin{equation}
\min_{\bm{\Xi}, \bm{\Lambda}, \lambda, \bm{\Gamma}} \left[ \|\dot{\mathbf{X}} - \bm{\Theta}(\mathbf{X}; \bm{\Lambda}) \bm{\Xi}\|_2^2 + \lambda \| \;|\bm{\Gamma}|\; \bm{\Xi}\|_1 \right] \;,
\label{optim_eq_adam_joint}
\end{equation}
where all parameters, including coefficients and sparsity weights, are optimized simultaneously by minimizing the objective function. 
In a first assessment, it appears that the joint minimization leads to \(\bm{\Gamma} \rightarrow 0\), effectively removing the \(\ell_1\) penalty from the loss function and allowing no sparsity in the model output. However, this progression towards zero is counteracted by sequential thresholding applied to both \(\bm{\Xi}\) and \(\bm{\Lambda}\). Thresholding out candidate functions with negligible contributions forces the corresponding \(\bm{\Gamma}\) values to saturate at nonzero levels, while the \(\bm{\Gamma}\) associated with significant candidates continues to decrease towards zero. The benefits of this are twofold. First, saturation of \(\bm{\Gamma}\) values limits the ability of undesired candidate functions to re-enter the model in later epochs. Second, free movement of \(\bm{\Gamma}\) towards zero for remaining candidate terms increases the relative influence of the \(\ell_2\) data-fidelity loss over the \(\ell_1\) sparsity penalty, encouraging higher model accuracy as convergence approaches the true candidate functions. The theoretical foundations of this phenomenon, alongside empirical evidence, are discussed further in the~\ref{appendix:comp}. We utilize joint-minimization for the main results in the paper and discuss min-max in~\ref{appendix:comp}.
}

In practice, either \(\lambda\) or \(\bm{\Gamma}\) is selected to enforce sparsity. Among these, \(\bm{\Gamma}\) is preferred due to its ability to treat each candidate function individually. Iteratively updating the weighting matrix \(\bm{\Gamma}\) shares similar strategies with the Iteratively Reweighted Least Squares (IRLS) technique, which is applied to both constrained~\cite{candes2008enhancing} and unconstrained  problems~\cite{lai2013improved}. In the present work, we adopt a similar strategy by iteratively updating the \(\bm{\Gamma}\) matrix, but with the distinction of updating the weights using gradients computed through backpropagation.

The incorporation of stochastic gradient descent into the ADAM-SINDy methodology within the PyTorch environment offers several notable advantages over the classical SINDy approach:

\begin{enumerate}
    \item \textbf{Simultaneous optimization:} ADAM-SINDy concurrently optimizes hyperparameters and identifies both nonlinear parameters and relevant candidate functions. Unlike SINDy, the proposed method does not require any prior knowledge about the nonlinear parameters in the candidate functions. ADAM-SINDy's integrated optimization framework reduces the need for extensive tuning processes, which are often required in traditional SINDy via trial-and-error methods or sequential optimization steps. 
    
    \item \textbf{Adaptive weighting:} ADAM-SINDy treats the regularization parameter, $\bm{\Gamma}$, as an optimizable matrix (inspired by the IRLS methodology), where the candidate-specific weights are also optimized over epochs. As a result, the method facilitates an individualized treatment of the candidate functions, thereby enhancing the accuracy of parameter identification and improving the overall robustness of the system identification process. \edit{We also show that this approach facilitates achieving machine-precision accuracy.}
    
    \item \textbf{Higher-order regularization:} Utilizing a computational graph, ADAM-SINDy further enables the opportunity to efficiently apply higher-order Tikhonov regularization, by incorporating a differential equation within the \(\ell_1\) norm, efficiently computed using autodifferentiation functionality. This method is inspired by the physics-based loss functions used in Physics-Informed Neural Networks (PINN)~\cite{raissi2019physics}, allowing for the imposition of complex constraints that ensure the identified model aligns closely with the governing physical laws. This augmented regularization framework guarantees that the identified dynamical system remains robust, interpretable, and consistent with the underlying physical principles. Here, we will briefly discuss this aspect and pursue the concept of physics-informed SINDy with ADAM-SINDy in future work.

\end{enumerate}

\edit{A detailed description of the ADAM-SINDy algorithm is provided in~\ref{appendix:alog}, accompanied by the discussion of both stochastic and full-batch optimization strategies. In~\ref{appendix:comp}, we further provide theoretical/empirical justification for why ADAM-SINDy's joint-minimization approach together with thresholding works and facilitates machine-precision accuracy.  }

\subsection{Problem Statement}
In this work, several benchmark dynamical systems problems are selected to illustrate the efficacy of the ADAM-SINDy framework and to underscore its advantages relative to the classical SINDy approach. All SINDy methodologies are built upon a common master library that incorporates a diverse range of nonlinear functions, including polynomials, Fourier terms, exponential functions, and nonlinear terms combined with polynomial expressions. This library is formally defined as follows:
\begin{equation}
\bm{\Theta} = \begin{bmatrix}
1 & \mathbf{X}^\mathcal{A} & \sin(\mathcal{B}\mathbf{X}) & \cos(\mathcal{C}\mathbf{X}) & \exp(\mathcal{D}\mathbf{X}) & \mathbf{X}\otimes\sin(\mathcal{E}\mathbf{X}) & \mathbf{X}\otimes\cos(\mathcal{F}\mathbf{X}) & \mathbf{X}\otimes \exp(\mathcal{G}\mathbf{X})
\end{bmatrix} \;,
\label{eq:candidate_library}
\end{equation}
where \(\mathbf{X}\) denotes the input data, and \(\mathcal{A}, \mathcal{B}, \mathcal{C}, \mathcal{D}, \mathcal{E}, \mathcal{F}, \mathcal{G}\) represent the adaptive nonlinear parameters \((\bm{\Lambda})\) that are subject to optimization within the ADAM-SINDy framework. Here, \(\mathcal{A}\) is the polynomial order, \(\mathcal{B}\) and \(\mathcal{C}\) represent Fourier frequencies, \(\mathcal{D}\) corresponds to exponential growth/decay, \(\mathcal{E}\) and \(\mathcal{F}\) are the Fourier frequencies for Fourier terms combined with polynomial terms, and \(\mathcal{G}\) represents the exponential growth/decay for the exponential terms combined with polynomial terms. Specifically, \(\mathcal{A}\) is chosen to be 1, thereby encompassing polynomial terms up to the first order in the master library for all cases. However, in certain numerical experiments, \(\mathcal{A}\) is rendered trainable when it corresponds to a non-integer rational exponent. It is noteworthy that the classical SINDy method relies on precise a priori knowledge of these parameters to successfully identify a sparse system.

The proposed ADAM-SINDy methodology also offers further significant advantages due to the trainable nature of the nonlinear parameters. Specifically, when the parameters associated with cosine and exponential functions—\(\mathcal{C}\) and \(\mathcal{D}\)—approach zero, their non-zero coefficients effectively interact with the constant term \(1\) present in the library. Similarly, in the context of nonlinear polynomial combined terms, specifically those involving exponential and cosine functions, when \(\mathcal{F}\) and \(\mathcal{G}\) converge to zero, these terms interact with the polynomial terms \(\mathbf{X}^\mathcal{A}\). This phenomenon of effective self-adaptation of nonlinear functions and its interaction with other functions, as distinctly observed in the ADAM-SINDy framework is particularly crucial in scenarios where predefined candidate functions lack an exact function required to accurately represent the true system dynamics, as the trainable parameters are optimized to compensate for these deficiencies.

In contrast, the classical SINDy method operates under the assumption of a convex loss landscape, where nonlinear parameters are generated by uniform sampling within a predefined range. This conventional approach not only restricts the flexibility of the parameters but also precludes interactions between candidate functions within the library, as evidenced in the ADAM-SINDy methodology. Consequently, the classical SINDy approach is inherently constrained by its reliance on static parameters, which significantly increases the likelihood of incorrect system identification, particularly when essential functions are absent from the candidate library. As a result, the classical SINDy method frequently struggles to accurately capture the true dynamics of the system, especially in scenarios with nonlinear parameters, wherein the rigidity of predefined parameters--coupled with the absence of precise values--further impedes the effective identification of governing equations.

To investigate these limitations, two variants of the classical SINDy approach are explored: the SINDy-W method and the SINDy-W/o method. The SINDy-W method represents a fortunate scenario (unlikely in practical complex problems) where the precise value of the unknown nonlinear parameter is included within the range of candidate parameters $\bm{\Lambda}$. While such a scenario may hold in controlled or well-defined environments, they are often inadequate in real-world engineering applications, where the underlying parameter values may be unknown or difficult to estimate. Conversely, the SINDy-W/o method assumes no prior knowledge of the parametric values and applies a broad range of distributed parameter values to account for a wider spectrum of possibilities. In our benchmark problems, the range of parameters is selected such that the unknown parameter falls within the range. The SINDy-W/o approach is the more realistic practical version of SINDy, however, it often struggles in cases when the predefined parameter ranges fail to contain the ``exact value" required to capture the system's true behavior. This absence of prior knowledge frequently leads to less accurate, overly complex, and non-parsimonious models. Ultimately, these challenges underscore the need for flexibility and adaptability in parameter selection to ensure accurate system identification, motivating the use of the ADAM-SINDy framework, which dynamically adjusts parameter values during optimization and overcomes the limitations of the classical methods' predefined parameter spaces.

Here, we adopt the approach established in the classical SINDy method regarding the sparsity-promoting parameters \(\lambda\) and \(\bm{\Gamma}\) ~\cite{brunton2016discovering}. Specifically, \(\bm{\Gamma}\) is treated as an identity matrix, with the \(\lambda\) parameter being the sole variable used to control sparsity for the two variants considered: classical SINDy-W and SINDy-W/o methods. Notably, the numerical experiments conducted with the ADAM-SINDy framework indicate that the simultaneous optimization and utilization of both sparsity-promoting knobs is not beneficial.



The ADAM-SINDy methodology is implemented using the PyTorch library, considering both batch-wise and full-batch optimization strategies, and employs the ADAM optimizer to facilitate the simultaneous optimization of all unknown variables involved in the sparse regression problem, including the parameters \(\bm{\Lambda}\), the coefficients \(\bm{\Xi}\), and the sparsity knob \(\bm{\Gamma}\), with the parameter \(\lambda\) fixed at a value of 1 and not subject to optimization. In this implementation, the parameter $\bm{\Lambda} \in \mathbb{R}^{m\times \hat{p}}$ is defined as a rectangular matrix associated with $\hat{p}$ nonlinear candidate parameters corresponding to each of the $m$ state variables. This structure facilitates vectorization, allowing the ADAM optimizer to be invoked only once for all nonlinear parameters rather than for each individual value. Consequently, while $\bm{\Lambda}$ parameters behave like scalar values algorithmically, it is implemented as a rectangular matrix in vectorized form for improved computational efficiency. Thus, the parameters $\bm{\Lambda}$ and $\bm{\Xi}$ are initialized as scalar matrices with all values of 1. Additionally, the diagonal matrix \(\bm{\Gamma}\), with trainable diagonal entries, is initialized using a Gaussian distribution with a standard deviation of 1. To enhance convergence during the optimization process, a step decay variant for the dynamic learning rate is employed, systematically reducing the learning rate with a decay rate of 0.5 across all numerical experiments conducted in this research. Furthermore, during the sequential thresholding step, a tolerance value of \(5 \times 10^{-3}\) is utilized for \(\epsilon\), effectively eliminating small \(\bm{\Xi}\) and \(\bm{\Lambda}\) values. A comprehensive overview of the hyperparameters for both the classical SINDy and ADAM-SINDy methods is provided in Table~\ref{table_1}, detailing the parametric values chosen for the comparative analysis.

\begin{table}[]
\begin{tabular}{|ccc|c|c|c|c|c|}
\hline
\multicolumn{3}{|c|}{Experiment} & \begin{tabular}[c]{@{}c@{}}Harmonic\\ Oscillator\end{tabular} & \begin{tabular}[c]{@{}c@{}}Van der pol\\ Oscillator\end{tabular} & ABC Flow & \begin{tabular}[c]{@{}c@{}}Chemical\\ Reaction\end{tabular} & \begin{tabular}[c]{@{}c@{}}Pharmacokinetics\\ Model\end{tabular} \\ \hline
\multicolumn{1}{|c|}{} & \multicolumn{2}{c|}{$\lambda$} & \begin{tabular}[c]{@{}c@{}}$ $\\ 0.01 \\ $ $\end{tabular} & 0.0075 & 0.1 & 0.025 & 0.01 \\ \cline{2-8} 
\multicolumn{1}{|c|}{} & \multicolumn{1}{c|}{\multirow{2}{*}{\begin{tabular}[c]{@{}c@{}}\\\\ \rotatebox[origin=c]{90}{SINDy - W}\end{tabular}}} & \multirow{2}{*}{\begin{tabular}[c]{@{}c@{}}$ $\\ \\ $\mathcal{A}$\end{tabular}} & \multirow{2}{*}{\begin{tabular}[c]{@{}c@{}}$ $\\ \\ \textbf{1.0}\end{tabular}} & \multirow{2}{*}{\begin{tabular}[c]{@{}c@{}}$ $\\ \\ \textbf{1.0}, \textbf{2.15}\end{tabular}} & \multirow{2}{*}{\begin{tabular}[c]{@{}c@{}}$ $\\ \\ 1.0\end{tabular}} & \multirow{2}{*}{\textbf{\begin{tabular}[c]{@{}c@{}}$ $\\ \\ 1.0\end{tabular}}} & \textbf{\begin{tabular}[c]{@{}c@{}}$ $\\ $\mathbf{X}^{\textbf{1.0}}$\\ $ $\end{tabular}} \\ \cline{8-8} 
\multicolumn{1}{|c|}{} & \multicolumn{1}{c|}{} &  &  &  &  &  & \begin{tabular}[c]{@{}c@{}}$ $\\ $t^{(-0.4,\textbf{-0.5}, -0.55)}$\\ $ $\end{tabular} \\ \cline{3-8} 
\multicolumn{1}{|c|}{} & \multicolumn{1}{c|}{} & $\mathcal{B},\mathcal{C},\mathcal{E},\mathcal{F}$ & \begin{tabular}[c]{@{}c@{}}0.25, 0.5,\\ \textbf{0.75}, 1.0\end{tabular} & $\pi$ & \begin{tabular}[c]{@{}c@{}}\boldmath$\frac{\pi}{2},\frac{\pi}{2.8},\frac{\pi}{3},$\\ \boldmath$\frac{\pi}{4},\frac{\pi}{4.5},\frac{\pi}{5}$\end{tabular} & $\pi$ & \begin{tabular}[c]{@{}c@{}}$ $\\ $\pi$\\ $ $\end{tabular} \\ \cline{3-8} 
\multicolumn{1}{|c|}{\rotatebox[origin=c]{90}{SINDy}} & \multicolumn{1}{c|}{} & $\mathcal{D},\mathcal{G}$ & \begin{tabular}[c]{@{}c@{}}$ $\\ 1.0\\ $ $\end{tabular} & 1.0 & 1.0 & \begin{tabular}[c]{@{}c@{}}1.0, \textbf{1.015} \\ \textbf{1.025}, 2\end{tabular} & 1.0, 2.0 \\ \cline{2-8} 
\multicolumn{1}{|c|}{} & \multicolumn{1}{c|}{\multirow{2}{*}{\begin{tabular}[c]{@{}c@{}}\\\\ {\rotatebox[origin=c]{90}{SINDy - W/o}}\end{tabular}}} & \multirow{2}{*}{\begin{tabular}[c]{@{}c@{}}$ $\\ \\ $\mathcal{A}$\end{tabular}} & \multirow{2}{*}{\begin{tabular}[c]{@{}c@{}}$ $\\ \\ 1.0\\ $ $\end{tabular}} & \multirow{2}{*}{\begin{tabular}[c]{@{}c@{}}$ $\\ \\ 1.0, 2.5\end{tabular}} & \multirow{2}{*}{\begin{tabular}[c]{@{}c@{}}$ $\\ \\ 1.0\end{tabular}} & \multirow{2}{*}{\begin{tabular}[c]{@{}c@{}}$ $ \\ \\ 1.0\end{tabular}} & \begin{tabular}[c]{@{}c@{}}$ $\\ $\mathbf{X}^{1.0}$\\ $ $\end{tabular} \\ \cline{8-8} 
\multicolumn{1}{|c|}{} & \multicolumn{1}{c|}{} &  &  &  &  &  & \begin{tabular}[c]{@{}c@{}}$ $\\ $t^{(-0.4, -0.55)}$\\ $ $\end{tabular} \\ \cline{3-8} 
\multicolumn{1}{|c|}{} & \multicolumn{1}{c|}{} & $\mathcal{B},\mathcal{C},\mathcal{E},\mathcal{F}$ & \begin{tabular}[c]{@{}c@{}}0.2, 0.5,\\ 0.7, 1.0\end{tabular} & $\pi$ & \begin{tabular}[c]{@{}c@{}}$\frac{\pi}{2}, \frac{\pi}{2.5},\frac{\pi}{3.5},$\\ $\frac{\pi}{4},\frac{\pi}{4.2},\frac{\pi}{5}$\end{tabular} & $\pi$ & \begin{tabular}[c]{@{}c@{}}$ $\\ $\pi$\\ $ $\end{tabular} \\ \cline{3-8} 
\multicolumn{1}{|c|}{} & \multicolumn{1}{c|}{} & $\mathcal{D},\mathcal{G}$ & \begin{tabular}[c]{@{}c@{}}$ $\\ 1.0\\ $ $\end{tabular} & 1.0 & 1.0 & 1.0, 2.0 & 1.0, 2.0 \\ \hline
\multicolumn{1}{|c|}{\multirow{3}{*}{\rotatebox[origin=c]{90}{ADAM-SINDy}}} & \multicolumn{2}{c|}{Epochs} & \begin{tabular}[c]{@{}c@{}}$$\\ 50000\\ $$\end{tabular} & 32000 & 60000 & 40000 & 40000 \\ \cline{2-8} 
\multicolumn{1}{|c|}{} & \multicolumn{2}{c|}{Initial learning rate} & \begin{tabular}[c]{@{}c@{}}$$\\ 0.1\\ $$\end{tabular} & 0.01 & 0.1 & 0.01 & 0.01 \\ \cline{2-8} 
\multicolumn{1}{|c|}{} & \multicolumn{2}{c|}{Step decay epochs} & \begin{tabular}[c]{@{}c@{}}$$\\ 4000\\ $$\end{tabular} & 2500 & 4000 & 3000 & 4500 \\ \hline
\end{tabular}
\caption{Hyperparameters utilized in the classical SINDy methods (SINDy-W and SINDy-W/o) and the ADAM-SINDy method across the benchmark problems considered in this study. The nonlinear parameters \(\mathcal{A}, \mathcal{B}, \mathcal{C}, \mathcal{D}, \mathcal{E}, \mathcal{F}, \mathcal{G}\) represent values employed in the library for the SINDy-W and SINDy-W/o approaches, demonstrating performance when the distributed range covers exact (SINDy-W) and approximate (SINDy-W/o) parameters of the system dynamics. Parameters in \textbf{bold} indicate exact values derived from the ground truth data in the SINDy-W method. For the pharmacokinetics model, \(\mathcal{A}\) refers to the exponents used for the state variable and the time-dependent function, enumerated accordingly. Hyperparameters specific to the wildfire PDE example are discussed in the relevant section.}
\label{table_1}
\end{table}

\section{Numerical Experiments}
\label{sec3}
This section details the benchmark problems used in this study, encompassing five systems of nonlinear ordinary differential equations (ODEs) and a sixth case involving a nonlinear partial differential equation (PDE). The final test case highlights the advantages of integrating trainable sparsifying hyperparameters, namely $\lambda$ and $\Gamma$, along with a sensitivity analysis. Each benchmark problem features a nonlinear parameter that poses challenges for the classical SINDy approach, and the SINDy-W method successfully identifies the governing dynamics for all cases as expected. These results underscore the challenges often encountered in practical SINDy applications (SINDy-W/o) when the correct nonlinear parameter is absent from the candidate library. 

\edit{Given the two objective formulations considered in this work--- joint minimization and min-max optimization---a comparative analysis presented in ~\ref{appendix:comp} demonstrates that the min-max approach is unsuitable due to the absence of a saddle point. This deficiency hinders convergence and requires further enhancements, such as weight decay, making the approach ultimately less robust. However, theoretical and empirical evidence support the joint minimization formulation as a viable strategy for the identification of nonlinear dynamical systems. Thus, the joint minimization approach has been adopted for the present work.}
In addition, the results from ADAM-SINDy using batch-wise optimization are excluded, as this method failed to produce accurate outcomes for complex dynamical systems such as the pharmacokinetic model and wildfire PDE identification, where a batch size of approximately 70-80\% of the entire dataset was required, making it nearly equivalent to full-batch optimization. Consequently, only the full-batch ADAM-SINDy results are presented here, as this approach yielded reliable and accurate solutions for all test cases.
\subsection{Harmonic Oscillator — Fourier Frequency}

First, we investigate Fourier frequencies within the candidate matrix by considering the harmonic oscillator as a benchmark problem. This system models a particle experiencing a restoring force proportional to its displacement from equilibrium. The dynamics of the system are governed by the following set of ODEs:

\begin{equation}
\begin{aligned}
\dot{x} &= ay \;, \\
\dot{y} &= -bx + cy \cos(dx) \;,
\end{aligned}
\end{equation}
where \( x(t) \) represents the displacement and \( y(t) \) represents the velocity. The term \( bx \) corresponds to the linear restoring force as described by Hooke’s law, while the cosine term introduces nonlinear effects through the frequency parameter \( d \), which captures external factors such as damping or external driving forces. This example illustrates the effectiveness of the ADAM-SINDy framework in accurately computing the Fourier frequency \( d \), as well as identifying other system coefficients, without the need for correct predefined parameter distributions, a limitation inherent to the classical SINDy approach. In this case, the cosine frequency \( d \) is set to 0.75, a non-integer value, while the other parameters are chosen as \( a = 1 \), \( b = 1 \), and \( c = 0.1 \), to represent the ground truth dynamics of the system, which is generated up to $t=50$ with the initial condition of  $x_{0} =-2 $ and $y_{0} = 0$, for a time step of $0.01$.

\begin{figure}  
    \centering
    \includegraphics[width=0.8\textwidth,height=\textheight,keepaspectratio]{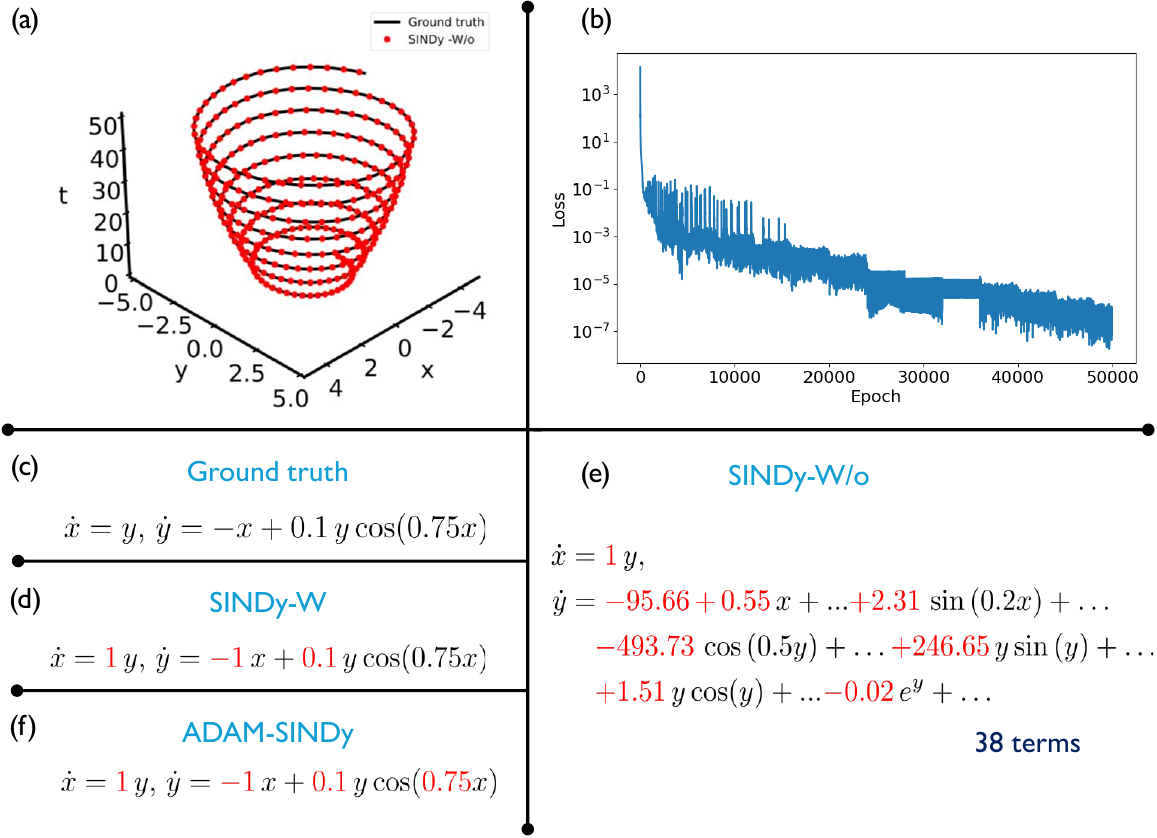}
    \caption{Harmonic oscillator example. (a) Temporal trajectory comparison showing both the ground truth and the SINDy-W/o method. (b) Loss plot illustrating the convergence behavior of the ADAM-SINDy method. A comparison of the governing equations is provided between the (c) ground truth and those identified by the (d) SINDy-W, (e) SINDy-W/o, and (f) ADAM-SINDy approaches. The red-highlighted numerical values indicate the optimized parameters obtained through these methodologies. For the SINDy-W/o method, a total of 38 terms were identified for $\dot{y}$, reflecting the complexity of the resulting dynamical system. Ground truth data is generated up to $t = 50$ with the initial conditions $x_{0} = -2$ and $y_{0} = 0$, using a time step of 0.01.}

    \label{fig: harmonic}
\end{figure}

The results, depicted in Fig.~\ref{fig: harmonic}, provide a comparative analysis of the governing equations corresponding to the ground truth system and those identified by the SINDy-W, SINDy-W/o, and ADAM-SINDy methodologies. While the classical SINDy-W/o method yielded a trajectory that closely matches the ground truth data, the identified system dynamics were unnecessarily complex and deviated from the true governing equations. In contrast, the ADAM-SINDy framework accurately computed the Fourier frequency and demonstrated a clear advantage by recovering the system’s dynamics with greater accuracy, even in the presence of nonlinearity, as evidenced by quantitative analysis. SINDy-W also captures the true dynamical system, albeit by assuming precise knowledge of the frequency \( d \) in its library. It should be noted that in all figures the trajectory for SINDy-W and ADAM-SINDy methods are not shown as these methods identify the exact governing equations, therefore the trajectories are identical to the ground-truth.

\subsection{Van der Pol Oscillator — Rational Exponent}
This example focuses on the computation of rational exponents within the candidate matrix by employing the Van der Pol oscillator, a standard example of a nonlinear dynamical system characterized by self-sustained oscillations. The system is governed by the following set of first-order differential equations 

\begin{equation}
\begin{aligned}
\dot{x} &= y \;, \\
\dot{y} &= \mu(1 - x^b)y - x \;,
\end{aligned}
\end{equation}
where the parameter \(\mu\) introduces nonlinear damping, thereby influencing both the amplitude and frequency of the oscillations. For small values of \(\mu\), the system behaves similarly to a simple harmonic oscillator. However, as \(\mu\) increases, the system's nonlinearity becomes more pronounced, leading to complex dynamics, including limit cycles with variable amplitude and frequency. The primary objective in this example is for the ADAM-SINDy framework to accurately identify the exponent \(b\) in the nonlinear term, particularly when \(b\) takes on rational values, as opposed to the conventional integer value of 2. Accurately identifying such rational exponents poses significant challenges for classical SINDy methods, which often struggle with precise computation in these cases (unless the precise exponent is known in SINDy-W). The ground truth for this system is defined by choosing \(\mu = 0.01\) and \(b = 2.15\), with the non-integer exponent contributing to the system’s complex nonlinear behavior, which is generated up to $t=200$ with the initial condition of  $x_{0} =0 $ and $y_{0} = -1$, for a time step of $0.01$. In addition to the first-order polynomials in the candidate master library, the $\mathcal{A}$ parameter in Eq.~\ref{eq:candidate_library} is made trainable in the ADAM-SINDy method to accommodate the chosen non-integer value, while a range of parameters $(1, 2.15)$ and $(1, 2.5)$ are selected for the classical SINDy-W and SINDy-W/o methodologies, respectively.

\begin{figure}  
    \centering
    \includegraphics[width=0.8\textwidth,height=\textheight,keepaspectratio]{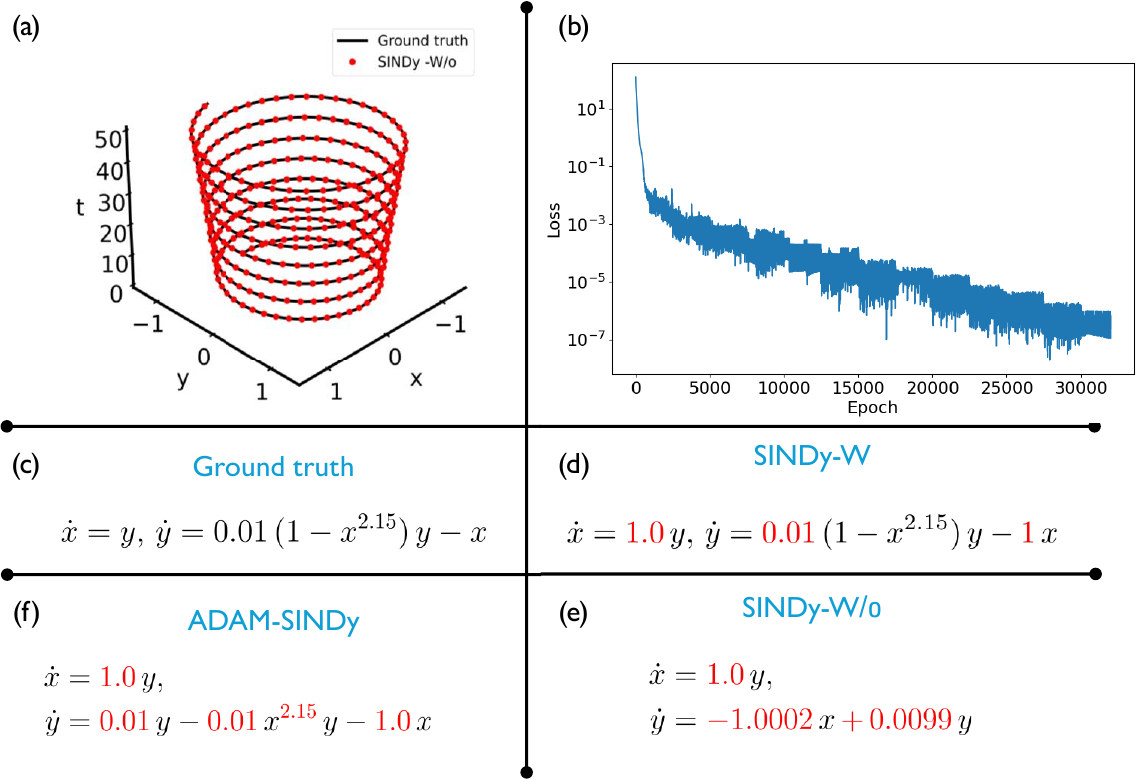}
    \caption{Van der Pol oscillator example. (a) Temporal trajectory comparison showing both the ground truth and the SINDy-W/o method. (b) Loss plot illustrating the convergence behavior of the ADAM-SINDy method. A comparison of the governing equations is provided between the (c) ground truth and those identified by the (d) SINDy-W, (e) SINDy-W/o, and (f) ADAM-SINDy approaches. The red-highlighted numerical values indicate the optimized parameters obtained through these methodologies. Ground truth data is generated up to $t = 200$ with the initial conditions $x_{0} = 0$ and $y_{0} = -1$, using a time step of 0.01.}

    \label{fig:vander}
\end{figure}

Fig.~\ref{fig:vander} presents a detailed comparison of the governing equations for the ground truth system and those identified by the SINDy-W, SINDy-W/o, and ADAM-SINDy methods. Although the SINDy-W/o method is able to align the trajectory with the ground truth data, it identifies a significantly more complex and dense equation compared to the original system. In contrast, the ADAM-SINDy framework demonstrates its ability to accurately compute the non-integer exponent \(b\) and other coefficients of the governing equations by self-adapting the exponent. This capability is particularly noteworthy, as it enables the ADAM-SINDy approach to correctly identify relevant terms within the system's governing equations, even when the exponent is a rational number. Consequently, the ADAM-SINDy framework provides a more parsimonious representation of the system dynamics, highlighting its efficacy in dealing with non-integer exponent values that are difficult to guess a priori in the library of candidate terms.

\subsection{ABC Flow — Multiple Fourier Frequencies}
In this section, we investigate the computation of multiple Fourier frequencies and amplitudes, focusing on the complex dynamics of an analytically defined chaotic fluid flow system. To facilitate this analysis, we examine the well-established benchmark problem known as the ABC (Arnold-Beltrami-Childress) flow, which is frequently employed to study intricate fluid dynamics and chaotic behavior in three-dimensional systems~\cite{oettinger2016autonomous}. The velocity components of the ABC flow, which adhere to incompressibility, are formulated as follows:

\begin{equation}
\begin{aligned}
\dot{x} &= A \sin\left(w_1 z\right) + C \cos\left(w_2 y\right) \;, \\
\dot{y} &= B \sin\left(w_3 x\right) + A \cos\left(w_4 z\right) \;, \\
\dot{z} &= C \sin\left(w_5 y\right) + B \cos\left(w_6 x\right) \;,
\end{aligned}
\end{equation}
where \(x\), \(y\), and \(z\) represent the spatial coordinates, \(A\), \(B\), and \(C\) denote the amplitudes, and \(w_i\) correspond to the frequencies.

\begin{figure}  
    \centering
    \includegraphics[width=\textwidth,height=\textheight,keepaspectratio]{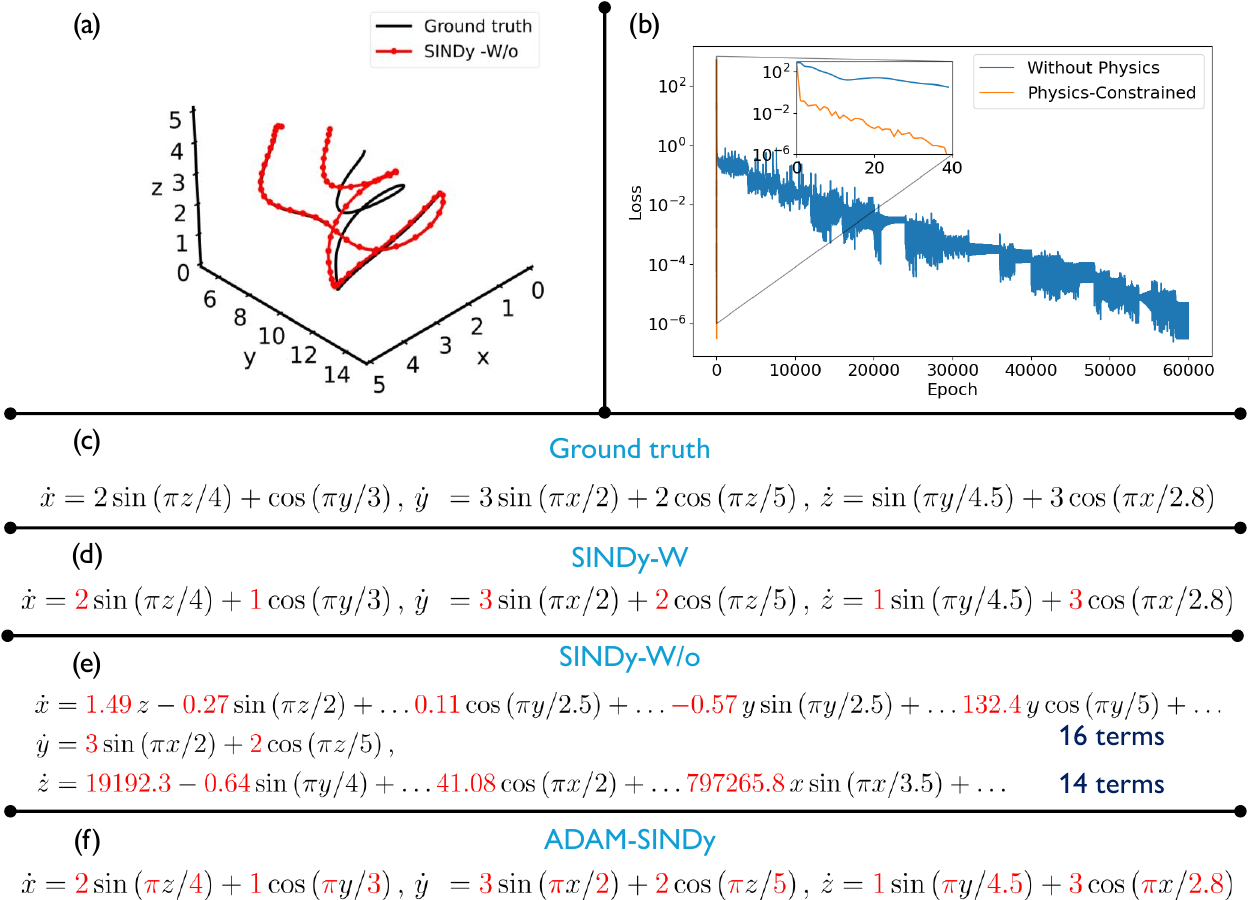}
    \caption{ABC flow example. (a) Pathline trajectory comparison showing both the ground truth and the SINDy-W/o method. (b) Loss plot illustrating the convergence behavior of the ADAM-SINDy method with and without divergence-free condition. A comparison of the governing equations is provided between the (c) ground truth and those identified by the (d) SINDy-W, (e) SINDy-W/o, and (f) ADAM-SINDy approaches. Optimized parameter values, highlighted in red, indicate results obtained through these methodologies. For the SINDy-W/o method, a total of 16 and 14 terms were identified for $\dot{x}$ and $\dot{z}$, reflecting the complexity of the derived dynamical system. In the ADAM-SINDy method, all optimized Fourier frequency values are approximated to 16 decimal places, consistently aligning with the value of $\pi$. Ground truth data is generated up to $t = 20$ with the initial conditions $x_{0} = 0.5$, $y_{0} = 0.2$, and $z_{0} = 1.0$, using a time step of 0.01.}

    \label{fig:abc_flow}
\end{figure}

The ABC flow gives rise to dynamic behaviors that range from simple periodic motions to complex, chaotic patterns, contingent on the specific parameters selected. A key characteristic of the ABC flow is its ability to produce intricate three-dimensional trajectories, making it a suitable model for examining chaotic dynamics in an autonomous dynamical system. Traditionally, the ABC flow has been modeled using a single Fourier frequency for simplicity. However, this work extends the analysis to encompass six distinct frequencies $w \in(\frac{\pi}{2},\frac{\pi}{2.8},\frac{\pi}{3}, \frac{\pi}{4},\frac{\pi}{4.5},\frac{\pi}{5})$, with Fourier amplitudes of $A = 2$, $B=3$, and $C= 1$, for which the ground truth generated up to $t = 20$ with the initial conditions $x_{0} = 0.5$, $y_{0} = 0.2$, and $z_{0} = 1.0$, using a time step of 0.01. In this multi-frequency context, both the SINDy-W and SINDy-W/o methods require an expanded candidate library, rendering the identification process more computationally intensive. By contrast, the ADAM-SINDy framework efficiently identifies all the Fourier frequencies and their corresponding amplitudes as illustrated in Fig.~\ref{fig:abc_flow}. As anticipated, the SINDy-W method successfully identifies all six Fourier frequencies and amplitudes. However, the SINDy-W/o method generates a complex and dense dynamical system equation that diverges significantly from the ground truth data, failing to produce accurate results despite including three correct frequencies among the six if the remaining three frequencies are incorrect. This underscores the importance of optimizing nonlinear parameters, positioning ADAM-SINDy as a robust and effective tool for identifying intricate dynamical systems such as the multi-frequency ABC flow.

Additionally, a supplementary experiment was conducted to explore the potential of utilizing the computational graph inherent to stochastic gradient descent (PyTorch environment) to employ a differential divergence form in the \(\ell_1\)-norm regularization term. In fluid dynamics, ensuring a divergence-free velocity field, which aligns with the conservation of mass, is a critical constraint. To incorporate this physical principle into the optimization process, the sparse regression problem is redefined as follows
\begin{equation}
\min_{\bm{\Xi}, \bm{\Lambda}}  \left[ \|\dot{\mathbf{X}} - \bm{\Theta}(\mathbf{X}\textbf{;}\thinspace \bm{\Lambda}) \thinspace \bm{\Xi}\|_2^2 + \lambda \|\hat{\Gamma}\thinspace\bm{\Xi}\|_1 \right] \;,
\label{physics_adam}
\end{equation}
where \(\hat{\Gamma}\) represents the divergence equation, defined as 
\(\partial \dot{x}_m / \partial x_m\) (where \(\dot{x}_m = \bm{\Theta}(\mathbf{X}; \bm{\Lambda}_m) \bm{\Xi}_m\)), computed efficiently using PyTorch's autodifferentiation functionality. Notably, \(\hat{\Gamma}\) is now a scalar value, distinct from the \(\bm{\Gamma}\) matrix used previously. Consequently, the \(\ell_1\)-norm regularization term encapsulates both the divergence equation loss and sparsity (\(\lambda > 0\)), thereby guiding the total loss function. This approach selectively retains candidate functions that satisfy the divergence-free condition, thereby ensuring compliance with the continuity equation within the ADAM-SINDy optimization framework. By fixing $\lambda$ to $1$, this novel approach enables the identification of all six Fourier frequencies and their corresponding amplitudes within just 40 epochs, as shown in Fig.~\ref{fig:abc_flow}, marking a substantial reduction in computational effort compared to the standard ADAM-SINDy method, which typically requires around 60,000 epochs when \(\bm{\Gamma}\) is a weighting matrix. This preliminary investigation suggests the potential for a physics-informed SINDy approach, which diverges from the primary focus of the current study and will be pursued in future work.

\subsection{Chemical Reaction Kinetics — Exponential Exponent}\label{chem_example}
Next, the computation of an exponential exponent within a dynamical system is examined through a chemical reaction kinetics problem that employs an exponential approximation of the Arrhenius rate law, which characterizes the nonlinear relationship between reaction rates and temperature, particularly in the context of coupled first-order reactions~\cite{boddupalli2023symbolic}. When the temperature increase due to an exothermic reaction is negligible compared to the ambient temperature, the reaction rate can be approximated as exponentially dependent on this incremental temperature change. By appropriately scaling the governing equations for mass and energy, the system is expressed as follows~\cite{boddupalli2023symbolic}

\begin{equation}
\begin{aligned}
\dot{\alpha} &= -k \alpha e^{g\theta} + \mu \;, \\
\dot{\theta} &= \alpha e^{h\theta} - \theta \;,
\end{aligned}
\end{equation}
where \(\alpha(t)\) represents the concentration of an intermediate reactant, \(\theta(t)\) denotes the temperature rise, \(k\) is the reaction rate constant, and \(\mu\) corresponds to the reactant concentration. The model assumes that the states \(\theta\) and \(\alpha\) evolve on a faster time scale compared to the slowly changing reactant consumption \(\mu\), invoking the pseudo-stationary-state hypothesis~\cite{gray1990chemical}, thus \(\mu\) interacting as a constant source term. This formulation offers insights into the temporal evolution of reactant concentrations and temperature, enabling the analysis of reaction kinetics and the impact of external perturbations on the reaction system.

\begin{figure}  
    \centering
    \includegraphics[width=0.9\textwidth,height=\textheight,keepaspectratio]{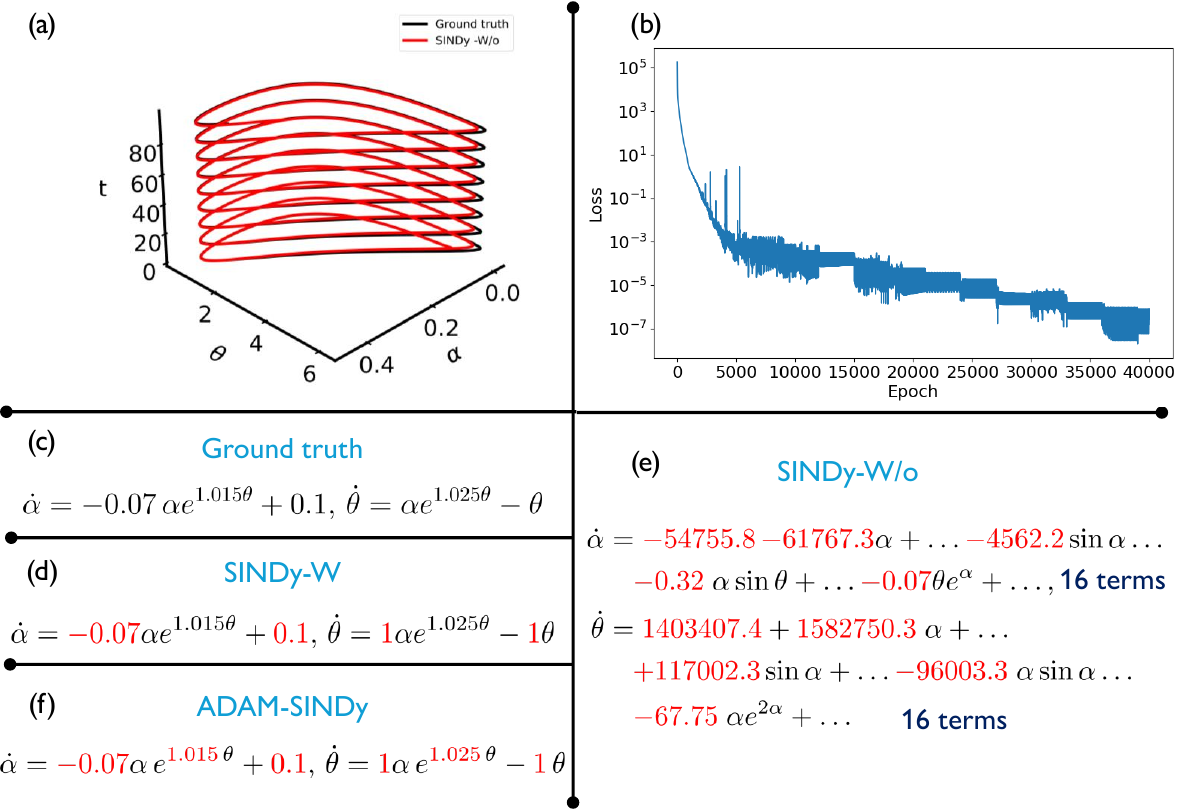}
    \caption{Chemical reaction kinetics example. (a) Temporal trajectory comparison showing both the ground truth and the SINDy-W/o method. (b) Loss plot demonstrating the convergence behavior of the ADAM-SINDy method. A comparison of the governing equations is provided between the (c) ground truth and those identified by the (d) SINDy-W, (e) SINDy-W/o, and (f) ADAM-SINDy approaches. The optimized parameters, highlighted in red, represent the results obtained from these methodologies. For the SINDy-W/o method, a total of 16 terms were identified for both $\dot{\alpha}$ and $\dot{\theta}$, indicating the complexity of the resulting dynamical system. Ground truth data is generated up to $t = 200$ with the initial conditions $\alpha_{0} = 0$ and $\theta_{0} = 5$, using a time step of 0.02.}

    \label{fig:chemical}
\end{figure}

Conventionally, the growth/decay rates \(g\) and \(h\) are assumed to be unity for the system under consideration~\cite{boddupalli2023symbolic, gray1990chemical}. However, this experiment adopts non-integer values for the parameters, specifically \( g = 1.015 \) and \( h = 1.025 \), while setting \( k = 0.07 \) and \( \mu = 0.1 \) to generate the ground truth data up to \( t = 200 \) with initial conditions \( \alpha_{0} = 0 \) and \( \theta_{0} = 5 \) using a time step of \( 0.02 \). Fig.~\ref{fig:chemical} provides a comprehensive comparison of the classical SINDy-W, SINDy-W/o, and ADAM-SINDy methods in identifying the chemical reaction system. In this context, the SINDy-W/o method exhibits deviation from the ground truth trajectory at $\alpha$ close to zero, failing to accurately replicate the dynamics. In contrast, the ADAM-SINDy method successfully identifies the non-integer exponent values and other relevant coefficients perfectly aligning with the ground truth data, thereby demonstrating its efficacy in addressing systems characterized by exponential functions.

\subsection{Pharmacokinetics Model — Time-dependent Reaction}
This example investigates a dynamical system that incorporates a time-dependent function with a non-integer exponent in its governing equations, focusing on a pharmacokinetics model. Such models are commonly utilized to assess the rate of drug concentration across various compartments in the human body. Specifically, we consider a single-dose compartmental model~\cite{ahmadi2024ai} governed by the following system of equations:

\begin{equation}
\begin{aligned}
k_g &= k_0 t^{-\eta} \;, \\
\dot{B} &= k_g G - k_b B \;, \\
\dot{G} &= -k_g G \;, \\
\dot{U} &= k_b B \;,
\end{aligned}
\end{equation}
where \(k_g\) denotes the time-dependent reaction rate~\cite{macheras2016modeling}, \(k_0\) is the initial rate coefficient, and \(\eta\) represents the fractal kinetics exponent, constrained within the range \(0 < \eta < 1\). The ground truth data are generated using a drug absorption rate of \( k_b = 0.15 \), $k_0 = 0.72$, and a fractal kinetics exponent of \( \eta = -0.5 \), evolved up to \( t = 10 \) with the initial conditions \( B_{0} = 0 \), \( G_{0} = 1 \), and \( U_{0} = 0 \) for a time step of \( 0.001 \).

\begin{figure}  
    \centering
    \includegraphics[width=1.0\textwidth,height=\textheight,keepaspectratio]{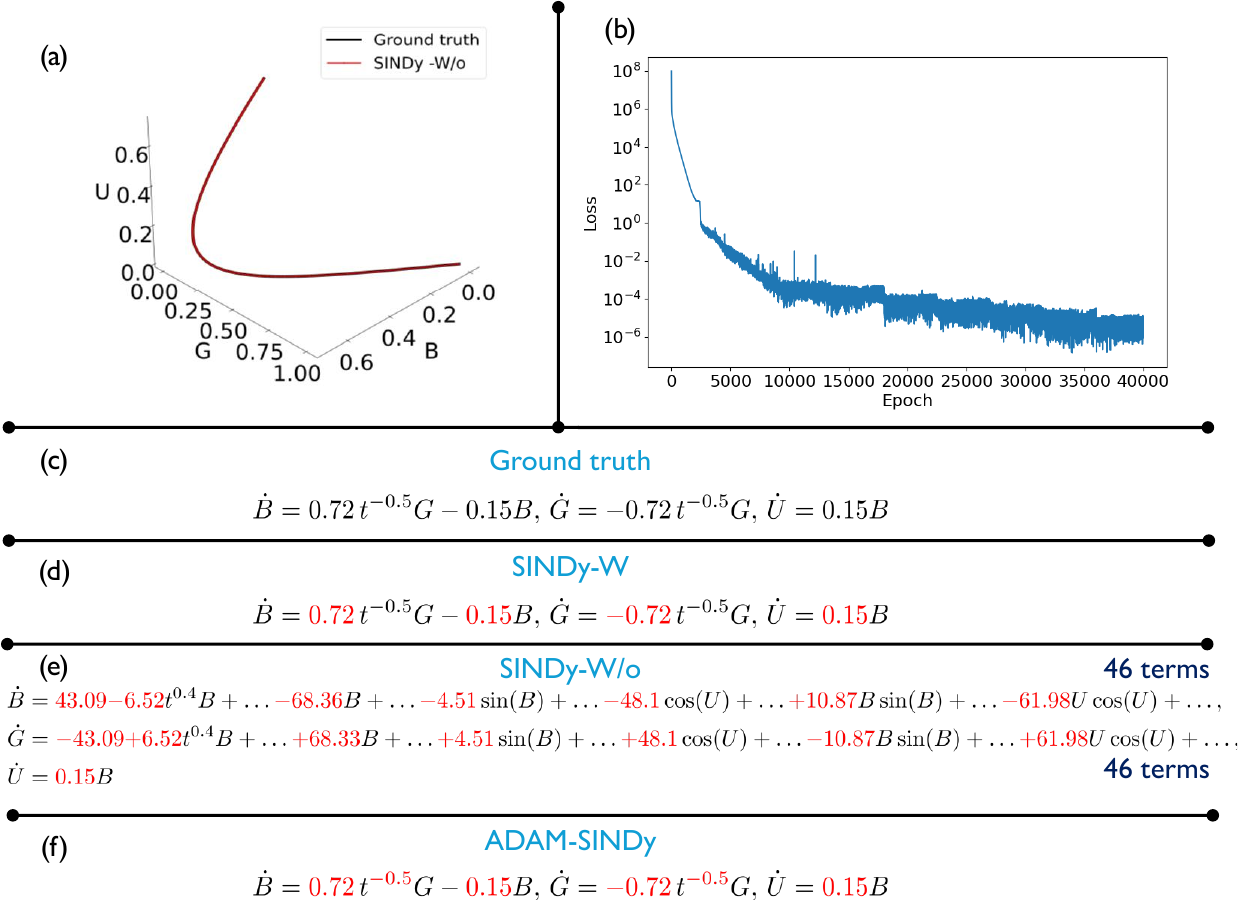}
    \caption{Pharmacokinetics example. (a) Temporal trajectory comparison highlighting both the ground truth and the SINDy-W/o method. (b) Loss plot illustrating the convergence behavior of the ADAM-SINDy method. A comparison of the governing equations is provided between the (c) ground truth and those identified by the (d) SINDy-W, (e) SINDy-W/o, and (f) ADAM-SINDy approaches. The optimized parameters, highlighted in red, reflect the outcomes of these methodologies. For the SINDy-W/o method, a total of 46 terms were identified for both $\dot{B}$ and $\dot{G}$, indicating the complexity of the resulting dynamical system. Ground truth data is generated up to $t = 10$ with the initial conditions $B_{0} = 0$, $G_{0} = 1$, and $U_{0} = 0$, using a time step of 0.001.}

    \label{fig:pharma}
\end{figure}

In this example, the master library is expanded to include \(\mathbf{X} \otimes t^{\eta}\), thereby accommodating the time-dependent function, with \(\eta\) treated as a trainable parameter within the ADAM-SINDy framework, while a range of parameters \((-0.4, -0.5, -0.55)\) and \((-0.4, -0.55)\) are defined for the SINDy-W and SINDy-W/o methodologies, respectively. Fig.~\ref{fig:pharma} presents a detailed comparison of the governing equations for the ground truth system with those identified by the SINDy-W, SINDy-W/o, and ADAM-SINDy methodologies. Notably, the SINDy-W/o method closely aligns with the ground truth trajectory, exhibiting no deviations. However, the resulting equation from this method is significantly more complex and dense compared to the original system. In contrast, the ADAM-SINDy framework accurately computes the exponent \(\eta\) associated with the temporal function, along with other coefficients, mirroring the governing system precisely and thereby demonstrating superior performance compared to the classical SINDy methods.

\subsection{Wildfire Transport - Partial Differential Equation}
The classical SINDy method was initially developed to discover systems of ODEs but has been extended to encompass PDEs through a process known as PDE-FIND~\cite{rudy2017data}. This extension is particularly pertinent for systems exhibiting dependencies in both spatial and temporal domains. In this section, we investigate wildfire dynamics modeled by PDEs, focusing on the identification of governing nonlinear parameters that elucidate the spatiotemporal evolution of temperature within the fire. These governing equations encapsulate critical processes, including heat transfer, combustion, and the influence of wind, thereby providing a comprehensive framework for understanding wildfire propagation across various fuel beds. The following non-dimensional wildfire combustion model~\cite{viknesh2024role} delineates this behavior, incorporating terms for advection, linear diffusion, reaction, and natural cooling:

\begin{align}
\frac{\partial T}{\partial t} + \overrightarrow{\mathbf{V}} \cdot \nabla T &= \kappa \nabla^2 T + \beta \mathrm{e}^{T/(1 + \epsilon T)} - \alpha T \;,
\end{align}
where \(T(\mathbf{x}, t)\) denotes the non-dimensionalized absolute temperature, \(\overrightarrow{\mathbf{V}}\) represents the normalized velocity vector of the fire front, \(\kappa\) is the normalized thermal diffusivity, \(\beta\) denotes the fuel mass fraction, \(\epsilon\) is the inverse activation energy, and \(\alpha\) represents the natural heat transfer coefficient. For this experiment, we consider the candidate matrix 

\begin{equation}
\bm{\Theta} = \begin{bmatrix}
1 & \mathbf{X} & \nabla \mathbf{X} & \nabla^2 \mathbf{X} & \nabla^3 \mathbf{X} & \nabla^4 \mathbf{X} & \exp\left(\frac{\mathbf{X}}{1+\mathcal{H}\mathbf{X}}\right)
\end{bmatrix} \;,    
\end{equation}
which incorporates a diverse range of candidate functions. This matrix includes polynomials up to the first order, as well as an exponential term with \(\mathcal{H}\) serving as an unknown optimizable parameter, consistent with Arrhenius' law—typically employed in modeling fuel combustion processes. Additionally, it encompasses higher-order partial spatial derivatives, extending to the fourth derivative.

\begin{figure}  
    \centering
    \includegraphics[width=0.9\textwidth,height=\textheight,keepaspectratio]{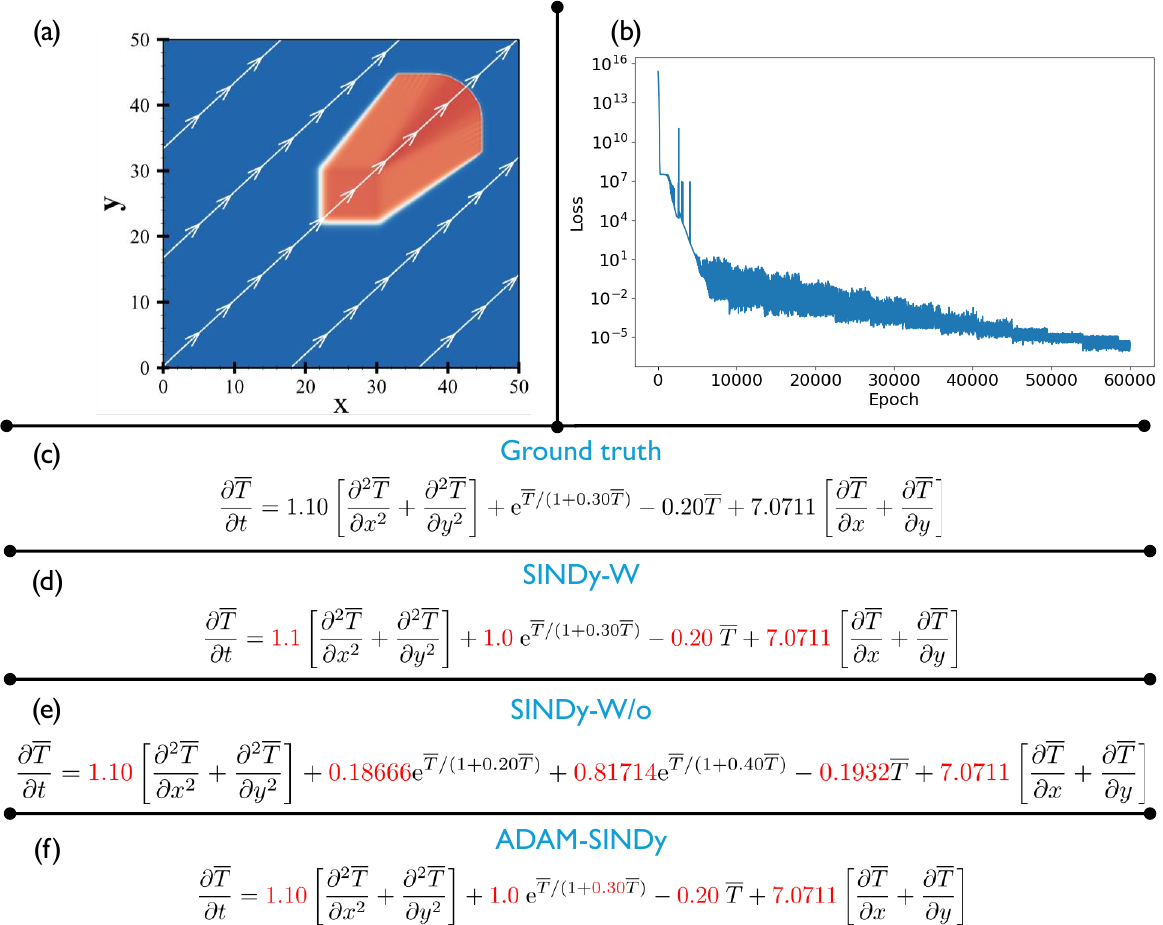}
    \caption{Wildfire dynamics example. (a) Ground truth temperature data superimposed with north-east facing wind streamlines at $t = 0.05$. (b) Loss plot illustrating the convergence behavior of the ADAM-SINDy method. A comparison of the governing equations is provided between the (c) ground truth and those identified by the (d) SINDy-W, (e) SINDy-W/o, and (f) ADAM-SINDy approaches. Optimized parameters, highlighted in red, indicate the results obtained through these methodologies. The initial Heaviside firefront is situated at coordinates $(8.0, 8.0)$ with a square side length of 4, computed up to $t = 0.08$ with a time step of $10^{-7}$.}

    \label{fig:wildfire}
\end{figure}

The ground truth data is generated by employing an upwind compact scheme for spatial derivatives and implicit-explicit Runge-Kutta time integration within the finite difference method, where the initial Heaviside firefront is situated at coordinates \((8.0, 8.0)\) with a square side length of 4, computed up to \( t = 0.08 \) with a time step of \( 10^{-7} \). The dimensionless parameters include a north-east facing freestream velocity magnitude of \(10\), a linear diffusion coefficient \(k = 1.1\), an inverse activation energy \(\epsilon = 0.3\), a fuel $\beta = 1$ and a natural heat transfer coefficient \(\alpha = 0.20\). In the context of the SINDy-W and SINDy-W/o methods, we consider the sets \(\mathcal{H} \in \{0.2, 0.3, 0.4\}\) and \(\{0.2, 0.4\}\) for the identification of the wildfire equation, with the sparsity parameter set to \(\lambda = 0.01\). Figure~\ref{fig:wildfire} illustrates a detailed comparison of the governing equations for the ground truth system with those identified by the SINDy-W, SINDy-W/o, and ADAM-SINDy methodologies. As anticipated, the SINDy-W/o method identifies an overly complex and dense wildfire model in comparison to the ground truth system. Interestingly, the ADAM-SINDy framework accurately identifies the exact wildfire combustion model, perfectly predicting the inverse activation energy associated with the reaction term.

\subsection{Hyperparameter Robustness — Sparsity Knobs}
\label{sec_hyper}
In this section, we investigate the optimization efficacy of the ADAM-SINDy framework in system identification, focusing on its sensitivity to the sparsity-promoting hyperparameters, \(\bm{\Gamma}\) and \(\lambda\). The performance of ADAM-SINDy is compared with the ideal case, the SINDy-W method. First, the SINDy-W method is employed to systematically vary the \(\lambda\) parameter across one order of magnitude, ranging from smaller to larger values, in order to determine the bounds beyond which the model diverges significantly from the ground truth. Specifically, we identify the minimum \(\lambda\), below which the SINDy-W method generates an overly dense system of equations, and the maximum \(\lambda\), beyond which the model fails to capture meaningful dynamics. The computed bound values for the SINDy-W method are presented in Table~\ref{tab:sparsity}.

In the context of the ADAM-SINDy method, two scenarios are considered: $(i)$ \(\bm{\Gamma}\) is fixed to the value of 1, while \(\lambda\) is trainable, and $(ii)$ \(\bm{\Gamma}\) is trainable, with its initialization varied by varying the standard deviation (\(\bm{\Gamma}_\sigma\)) of a Gaussian distribution while keeping its mean (\(\bm{\Gamma}_\mu\)) at zero, and \(\lambda\) is fixed at a value of 1. Experiments are conducted to compute the bounds for both scenarios across all the numerical examples considered in the present work, and the corresponding bounds are tabulated in Table~\ref{tab:sparsity}. Generally, the ADAM-SINDy framework provides much wider bounds for the sparsity knobs, often at least one order magnitude higher, especially in cases with higher initialization values, indicating that ADAM-SINDy is more robust than the classical SINDy method. Comparing the trainable \(\lambda\) and trainable \(\bm{\Gamma}\) cases in ADAM-SINDy, it is observed that in each example one of the two approaches provides a wider bound for the sparsity knob, indicating no clear advnatage.  However, in the former scenario (trainable \(\lambda\)), ADAM-SINDy identifies the correct candidate terms but the coefficients and nonlinear parameters are not always machine precision. In the latter case (trainable \(\bm{\Gamma}\)), the ADAM-SINDy method identifies the exact correct terms and nonlinear parameter values with machine precision. 

\begin{figure}  
    \centering
    \includegraphics[width=0.9\textwidth,height=\textheight,keepaspectratio]{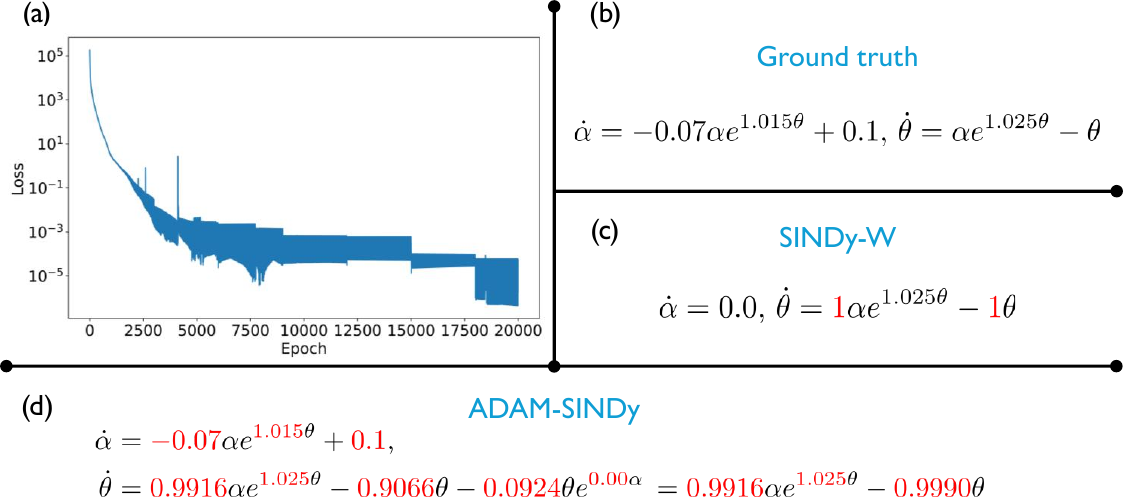}
    \caption{The influence of the sparsity knob ($\lambda$) on chemical reaction kinetics results. (a) Loss plot showing the convergence of the ADAM-SINDy method for $\lambda = 0.1$. A comparison of the governing equations is provided between the (b) ground truth and those identified by the (c) SINDy-W and (d) ADAM-SINDy methods, for $\lambda = 0.1$. The red-highlighted numerical values indicate the optimized parameters obtained.}

    \label{fig:lambda}
\end{figure}

Our experiments demonstrate that the latter optimization strategy, which imposes candidate-wise sparsity, is recommended for nonlinear system identification. A notable example is the chemical kinetics problem, illustrated in Fig.~\ref{fig:lambda}, which represents the former scenario of trainable \(\lambda\). It is observed that the SINDy-W method fails to produce the ground-truth equation, whereas ADAM-SINDy remains robust, producing the correct terms with slight deviations in the parameter values when initialized with a higher \(\lambda\) value of 0.1. Furthermore, results for the latter scenario with trainable \(\bm{\Gamma}\) are not shown here, as the identified terms and parameter values are observed to match the ground truth exactly, as demonstrated in Fig.~\ref{fig:chemical}.

\begin{table}[]
\centering
\begin{tabular}{|c|c|cc|}
\hline
\multirow{2}{*}{Experiment} & \multirow{2}{*}{\begin{tabular}[c]{@{}c@{}}$ $\\ SINDy-W\\ $ $\end{tabular}} & \multicolumn{2}{c|}{\begin{tabular}[c]{@{}c@{}}$ $\\ ADAM-SINDy\\ $ $\end{tabular}} \\ \cline{3-4} 
 &  & \multicolumn{1}{c|}{\begin{tabular}[c]{@{}c@{}}$ $\\ $\lambda$\\ $ $\end{tabular}} & \begin{tabular}[c]{@{}c@{}}$ $\\ $\bm{\Gamma}$\\ $ $\end{tabular} \\ \hline
\begin{tabular}[c]{@{}c@{}}$ $\\ Harmonic oscillator\end{tabular} & \begin{tabular}[c]{@{}c@{}}$ $\\ $10^{-3}$ $<$ $\lambda$ $<$ $10^{-1}$\\ $ $\end{tabular} & \multicolumn{1}{c|}{\begin{tabular}[c]{@{}c@{}}$ $\\ $10^{-3}$ $<$ $\lambda$ $<$ 10\\ $ $\end{tabular}} & \begin{tabular}[c]{@{}c@{}}$ $\\ $10^{-8}$ $<$ $\bm{\Gamma}_\sigma$ $<$ 10\\ $ $\end{tabular} \\ \hline
\begin{tabular}[c]{@{}c@{}}$ $\\ Van der pol oscillator\end{tabular} & \begin{tabular}[c]{@{}c@{}}$ $\\ $10^{-3}$ $<$ $\lambda$ $<$ $10^{-2}$\\ $ $\end{tabular} & \multicolumn{1}{c|}{\begin{tabular}[c]{@{}c@{}}$ $\\ $10^{-5}$ $<$ $\lambda$ $<$ 10\\ $ $\end{tabular}} & \begin{tabular}[c]{@{}c@{}}$ $\\ $10^{-5}$ $<$ $\bm{\Gamma}_\sigma$ $<$ 10\\ $ $\end{tabular} \\ \hline
\begin{tabular}[c]{@{}c@{}}$ $\\ ABC flow\\ $ $\end{tabular} & \begin{tabular}[c]{@{}c@{}}$ $ \\ $10^{-1}$\\ $ $\end{tabular} & \multicolumn{1}{c|}{\begin{tabular}[c]{@{}c@{}}$ $\\ $10^{-1}$\\ $ $\end{tabular}} & \begin{tabular}[c]{@{}c@{}}$ $\\ $10^{-2}$ $<$ $\bm{\Gamma}_\sigma$ $<$ 10\\ $ $\end{tabular} \\ \hline
\begin{tabular}[c]{@{}c@{}}$ $\\ Chemical reaction\end{tabular} & \begin{tabular}[c]{@{}c@{}}$ $\\ $10^{-3}$ $<$ $\lambda$ $<$ $10^{-1}$\\ $ $\end{tabular} & \multicolumn{1}{c|}{\begin{tabular}[c]{@{}c@{}}$ $\\ $10^{-12}$ $<$ $\lambda$ $<$ 10\\ $ $\end{tabular}} & \begin{tabular}[c]{@{}c@{}}$ $\\ $10^{-9}$ $<$ $\bm{\Gamma}_\sigma$ $<$ 10\\ $ $\end{tabular} \\ \hline
\begin{tabular}[c]{@{}c@{}}$ $\\ Pharmacokinetics\end{tabular} & \begin{tabular}[c]{@{}c@{}}$ $\\ $10^{-2}$ $<$ $\lambda$ $<$ $10^{-1}$\\ $ $\end{tabular} & \multicolumn{1}{c|}{\begin{tabular}[c]{@{}c@{}}$ $\\ $10^{-3}$ $<$ $\lambda$ $<$ 10\\ $ $\end{tabular}} & \begin{tabular}[c]{@{}c@{}}$ $\\ 1 $<$ $\bm{\Gamma}_\sigma$ $<$ 10\\ $ $\end{tabular} \\ \hline
\begin{tabular}[c]{@{}c@{}}$ $\\ Wildfire PDE\\ $ $\end{tabular} & \begin{tabular}[c]{@{}c@{}}$ $\\ $10^{-6}$ $<$ $\lambda$ $<$ 1\\ $ $\end{tabular} & \multicolumn{1}{c|}{-} & \begin{tabular}[c]{@{}c@{}}$ $\\ $10^{-6}$ $<$ $\bm{\Gamma}_\sigma$ $<$ 1\\ $ $\end{tabular} \\ \hline
\end{tabular}
\caption{Bounds on the sparsity-promoting hyperparameters for the SINDy-W and ADAM-SINDy methodologies, illustrating the conditions under which each method accurately identifies the dynamical system. In the ADAM-SINDy method, $\lambda$ is trainable while $\bm{\Gamma}$ is fixed at 1 in the first column. It is noted that no $\lambda$ values were found to work for the wildfire PDE example. In the second column, $\bm{\Gamma}$ is trainable while $\lambda$ is fixed at 1 and $\bm{\Gamma}$ is initialized using a Gaussian distribution where its mean $\bm{\Gamma}_\mu$ is fixed at zero and its standard deviation $\bm{\Gamma}_\sigma$ is varied as shown.}
\label{tab:sparsity}
\end{table}

\section{Discussion}
\label{sec4}
Identification of nonlinear dynamical systems is a prominent research domain that plays a pivotal role in elucidating the underlying physics of various problems. The literature highlights extensive investigations that employ the SINDy and symbolic regression frameworks, both of which can identify governing equations from observational data. However, it is essential to recognize that these well-established methods possess distinct advantages and disadvantages, particularly concerning computational optimization. In this context, the newly developed ADAM-SINDy framework emerges at the intersection of these established approaches, leveraging their computational optimization benefits. By employing the ADAM optimization algorithm, ADAM-SINDy facilitates the simultaneous optimization of parameters and coefficients associated with candidate nonlinear functions. This innovative approach enables more efficient and accurate parameter estimation without the necessity for prior knowledge of nonlinear characteristics, such as frequencies in trigonometric functions or bandwidths in exponential functions. A key factor in ADAM-SINDy's success lies in its ability to exploit the computational graph inherent in the gradient descent algorithm implementation in the PyTorch environment. This structure facilitates efficient backpropagation of gradients, enabling dynamic interaction between candidate terms and nonlinear parameters during the optimization process. By allowing the model to adaptively refine both the nonlinear parameters and coefficients, ADAM-SINDy enhances the robustness and precision of system identification, particularly in complex dynamical systems where the classical SINDy method struggles with sparse, non-convex landscapes. Notably, the strategy behind the simultaneous optimization of all unknown parameters and coefficients closely parallels the use of adaptive activation functions~\cite{jagtap2020adaptive} in neural network research, where the parameters of nonlinear adaptive activation functions, along with the weights and biases of linear layers, are optimized concurrently.

Interestingly, the ADAM-SINDy framework also facilitates interaction among the functions within the candidate library through the self-adaptation of nonlinear functions. This self-adaptive mechanism significantly enhances the model's capacity to capture complex dynamical behaviors, thereby allowing it to adjust to the varying degrees of nonlinearity inherent in real-world systems. Such adaptability is particularly crucial when predefined candidates are lacking, as it optimizes parameters to compensate for absent functions. In contrast, the classical SINDy approach is constrained by static nonlinear parameters and a lack of interaction, increasing the risk of inaccurate system models when essential candidate functions are omitted. This adaptability parallels the advantages observed in symbolic regression techniques, further enriching the model’s applicability. However, the occurrence of self-adaptation may transform the candidate matrix prior to optimization into an ``ill-conditioned'' matrix following optimization. Based on our tests, ill-conditioned cases do not pose a problem for ADAM-SINDy and simply result in two identical terms (in different forms) that need to be manually added for simplification of the final equation.


\edit{Concurrently, ADAM-SINDy seeks to identify a parsimonious mathematical model while optimizing trainable sparsity-promoting hyperparameters, such as $\bm{\Gamma}$, through two distinct objective formulations: joint-minimization and min-max. While the  min-max adaptive weighting strategies also appear in frameworks like PINNs~\cite{mcclenny2023self} and Transformers~\cite{correia2019adaptively}, we show that the joint-minimization strategy is more effective in practice, primarily because the convex nature of the $\ell_1$ sparsity term over both $\bm{\Gamma}$ and $\bm{\Xi}$ precludes the existence of a saddle point required for the min-max formulation to converge reliably. As demonstrated in the sensitivity analysis (Sec.~\ref{sec_hyper}), both trainable hyperparameters $\bm{\Gamma}$ and $\lambda$ can be used for ADAM-SINDy optimization; however, prioritizing $\bm{\Gamma}$ is recommended, as it consistently recovers the correct terms with machine precision, whereas $\lambda$ may yield correct structures with slight numerical deviations. ADAM-SINDy supports both batch-wise and full-batch optimization. For simpler dynamical systems, small randomly sampled batches often capture sufficient temporal information to accurately estimate governing parameters. In contrast, complex systems may require larger batches to encompass the necessary information, making full-batch optimization preferable. Accordingly, full-batch training is adopted in this work, leading to reduced uncertainty in the converged parameters within the ADAM-SINDy framework.


Particular emphasis is placed on the essential steps required for the joint-minimization approach to succeed, given the counterintuitive nature of this optimization problem. First, $\bm{\Gamma}$ must be incorporated within the $\ell_1$ loss as a transformed value. In the present work, that transformation is the absolute value function, although alternatives such as an exponential functions are possible. Taking the absolute value of the $\bm{\Gamma}$ matrix prior to computing the loss forces $\bm{\Gamma} \to 0$ rather than $-\infty$. This absolute value function also introduces stochasticity into the trajectories of $\bm{\Gamma}$, as zero crossings occur, subsequently influencing $\bm{\Xi}$ and $\bm{\Lambda}$, despite the use of full-batch optimization in the examples. Second, sequential thresholding is required to select a sparse system within ADAM-SINDy. Without this step, all $\bm{\Gamma}$ values would tend toward zero, eventually reducing the problem to a least-squares fit after sufficient epochs. Thresholding allows for saturation of selected $\bm{\Gamma}$ values, thereby establishing a fixed minimum level of sparsity that can only increase with each epoch. The specific threshold value itself is an additional hyperparameter that may vary depending on the specific problem; however, the same threshold value was used across all examples in the present work, indicating robust behavior. The theoretical foundations of the joint-minimization approach are further elaborated in the~\ref{appendix:comp}.

}

Ultimately, the integrated simultaneous adaptive optimization of the nonlinear parameters, coefficients matrix, and hyperparameters based on training data positions ADAM-SINDy as a powerful and effective tool for modeling a diverse range of dynamical behaviors across various applications. The proposed ADAM-SINDy methodology is thoroughly investigated across a wide spectrum of dynamical systems, including benchmark coupled ODEs such as oscillators, fluid flows, reaction kinetics, and wildfire dynamics PDE. These comprehensive investigations demonstrated its efficacy in accurately computing nonlinear features such as multiple Fourier frequencies, non-integer exponents in polynomials, and exponential exponents. Furthermore, the investigation also explored the inherent limitations of the classical SINDy method, illustrating its consistent failure to identify the exact model when the precise parameters are absent from the candidate matrix. This comprehensive investigation highlights the considerable potential of the ADAM-SINDy optimization framework, thereby extending its applicability to the modeling of intricate and diverse dynamical systems across various contexts with limited prior knowledge.

Similarly, symbolic regression via neural networks has been employed to identify dynamical systems~\cite{boddupalli2023symbolic}. Boddupalli et al.~\cite{boddupalli2023symbolic} applied this approach to chemical reaction kinetics, the same system considered in this work (Sec.~\ref{chem_example}), resulting in an overly complex, ``bloated'' mathematical model when explicit stacking within the neural network is absent. However, when exponential terms are explicitly stacked with polynomials, the identified model closely approximates the ground truth equation but exhibits slight deviations in the coefficient values.  Remarkably, ADAM-SINDy accurately identified the true dynamical system, capturing the dynamics without requiring additional treatments to ensure regression convergence.

\edit{In real-world scenarios, observational data are inevitably contaminated by noise from various sources, motivating an evaluation of ADAM-SINDy’s efficacy under such conditions. To assess its robustness, we compared ADAM-SINDy with classical SINDy-W by introducing Gaussian white noise up to 50\% of the standard deviation across all numerical examples. Although the results of these experiments are not included here, both methods exhibited comparable performance and failed to show clear advantages at high noise levels. At lower noise levels, both remained effective, suggesting that neither method consistently outperforms the other in the presence of significant noise. Future work will aim to enhance ADAM-SINDy’s resilience to noisy data.

A notable limitation of the full batch-ADAM-SINDy methodology is its high computational cost, stemming from the use of the ADAM optimizer for the coefficient matrix. While it offers advantages over classical SINDy, this improvement comes with considerable overhead, despite mirroring the computational strategies of its predecessor apart from the optimization algorithm. Efficiency can potentially be improved by employing an iterative least squares method for the coefficient matrix while retaining ADAM for nonlinear parameters and sparsity knobs, forming a combined least-squares–ADAM framework. However, integrating these methods poses challenges, particularly in aligning sparsity constraints with nonlinear optimization objectives. Further research is needed to develop robust methodologies that effectively merge these approaches while preserving their respective strengths. Notably, the ABC flow example demonstrated the benefit of embedding physical constraints directly into the learning process: incorporating a divergence-free condition into the loss function via the computational graph led to a dramatic reduction in training time from approximately 60,000 to just 40 epochs. This result highlights how leveraging underlying physical laws can not only enhance model fidelity but also significantly reduce computational cost, offering a promising direction for future work.

Given the primary focus on ``concurrent nonlinear parameter computations'' to address a key limitation of classical SINDy, ADAM-SINDy still does not fully overcome the challenge of constructing a master library containing all possible candidate terms for the numerical examples considered herein. Consequently, although a master library is included in the present problem setup, additional candidate functions were incorporated as needed for some specific problems. It is important to emphasize that this limitation is inherent to this class of problems. Classical SINDy acknowledged this constraint (see supplementary material in~\cite{brunton2016discovering}) through an illustrative example involving sine series and polynomial summations. This example demonstrated that nonlinear functions tend to approximate higher-order polynomials as sparsity is enforced. Likewise, symbolic regression also begins the system identification with simple polynomial functions and proceeds through the evolution stage, progressively introducing higher-order polynomials and nonlinear terms. Thus, the development of new methodologies is necessary to overcome the difficulty of accurately identifying the correct candidate functions, despite the presence of high-order polynomials and complex nonlinearities.

}

\section{Conclusion}
\label{sec5}
In summary, the proposed ADAM-SINDy methodology successfully addresses the limitations of the classical SINDy framework by enabling simultaneous identification of both system nonlinear parameters and coefficients through ADAM optimization. By eliminating the need for predefined parameter distributions and extensive manual hyperparameter tuning, ADAM-SINDy enhances the efficiency, adaptability, and robustness of system identification, particularly in the presence of nonlinear parameters. The demonstrated improvements across diverse dynamical systems highlight the potential of this approach to significantly broaden the applicability of the SINDy framework in complex modeling challenges.

\section*{Conflict of Interest}
The authors declare no conflict of interest.

\section*{Acknowledgments}
This research was supported by the National Science Foundation under Award No. 2247173 and 2330212. The authors express their gratitude to Dr. Jacob Hochhalter, Dr. Sarang Joshi, and Dr. Mike Kirby for their valuable discussions about this work.

\section*{Data Availability}
The ADAM-SINDy framework, developed in the PyTorch environment, is publicly available at \url{https://github.com/siva-viknesh/ADAM-SINDy}.

\appendix

\section{ADAM-SINDy Optimization Algorithm}\label{appendix:alog}
The ADAM-SINDy framework, detailed in Algorithm \ref{alg:ADAM-SINDy}, introduces a fully adaptive global optimization process for all parameters typically preselected in the classical SINDy framework. It begins with the initialization of key parameters—$\bm{\Xi}$, $\bm{\Lambda}$, $\lambda$, and $\bm{\Gamma}$—which are iteratively optimized based on the provided input data. To enhance computational efficiency and robustness, the input data is divided into batches of size $B$. When the batch size is set equal to the size of the input data, the algorithm operates under a full-batch optimization procedure. During each epoch, for each batch (or the full dataset in the case of full-batch optimization), the loss function $\mathcal{L}_b$ is computed to measure the discrepancy between the observed system dynamics, $\dot{\mathbf{X}}_b$, and the model’s predicted dynamics, represented by $\bm{\Theta}(\mathbf{X}_b; \bm{\Lambda})\thinspace\bm{\Xi}$. This batch-wise loss computation incorporates a sparsity-inducing term, regulated by either $\lambda$ or $\bm{\Gamma}$ knob, ensuring that only the most relevant terms are selected.

The parameters \(\bm{\Xi}\), \(\bm{\Lambda}\), \(\lambda\), and \(\bm{\Gamma}\) are then updated using stochastic gradient descent on a per-batch basis, progressively minimizing the loss function over the entire dataset. For full-batch optimization, updates are applied after evaluating the loss on the complete dataset. \edit{Importantly, the updates for the regularization parameters \(\lambda\) and \(\bm{\Gamma}\) differ from those of \(\bm{\Xi}\) and \(\bm{\Lambda}\). Depending on the optimization formulation, the regularization parameters are updated via either gradient ascent or gradient descent, governed by a sign variable \(s\). Specifically, in the min-max optimization formulation of Eq.~\eqref{optim_eq_adam}, \(\lambda\) and \(\bm{\Gamma}\) are updated through gradient ascent (\(s = +1\)), whereas in the joint minimization formulation of Eq.~\eqref{optim_eq_adam_joint}, the updates proceed via gradient descent (\(s = -1\)). This flexible update methodology enables adaptive control of sparsity throughout training.}
After each batch update, a sequential thresholding step sets entries of both \(\bm{\Xi}\) and \(\bm{\Lambda}\) to zero if their magnitudes fall below a specified tolerance \(\epsilon\). This iterative procedure continues for each batch until the maximum number of iterations \(N_{\text{max}}\) is reached, ensuring efficient convergence to a sparse and accurate model representation.

\begin{algorithm}
\caption{\textbf{ADAM-SINDy Algorithm}}
\label{alg:ADAM-SINDy}
\begin{algorithmic}
    \State \textbf{Input:}
    \State \quad Observational data $\mathbf{X}$ and $\dot{\mathbf{X}}$
    \State \quad Tolerance $\epsilon$ and maximum iterations $N_{\text{max}}$
    \State \quad Learning rate $\eta_{\bm{\Xi}}$, $\eta_{\bm{\Lambda}}$, $\eta_{\lambda}$, and $\eta_{\bm{\Gamma}}$ 
    
    \State \textbf{Initialization:}
    \State \quad Initialize $\bm{\Xi} \gets \bm{\Xi}_0$, $\bm{\Lambda} \gets \bm{\Lambda}_0$, $\lambda \gets \lambda_0$, $\bm{\Gamma} \gets \bm{\Gamma}_0$
    \State \quad Set iteration counter $n \gets 0$
    
    \State \textbf{Optimization Loop:}
    \While{$n < N_{\text{max}}$}
        \State \quad \textbf{For each batch} $b = 1, 2, \dots$
        \State \quad \quad Sample a batch of data $\mathbf{X}_b$ and $\dot{\mathbf{X}}_b$
        \State \quad \quad Compute $\bm{\Theta}(\mathbf{X}_b; \bm{\Lambda})$
        
        \State \quad \quad \textbf{Compute Loss Function:}  
        \[
        \mathcal{L}_b \gets \left\| \dot{\mathbf{X}}_b - \bm{\Theta}(\mathbf{X}_b; \bm{\Lambda}) \bm{\Xi} \right\|_2^2 + \lambda \left\| |\bm{\Gamma}| \bm{\Xi} \right\|_1
        \]
        
        \State \quad \quad \textbf{Update Parameters using Backpropagation:}
        \[
        \bm{\Xi}_{n+1} \gets \bm{\Xi}_n - \eta_{\bm{\Xi}} \nabla_{\bm{\Xi}} \mathcal{L}_b(\bm{\Theta}_b, \bm{\Xi}, \bm{\Lambda}, \lambda, \bm{\Gamma})
        \]
        \[
        \bm{\Lambda}_{n+1} \gets \bm{\Lambda}_n - \eta_{\bm{\Lambda}} \nabla_{\bm{\Lambda}} \mathcal{L}_b(\bm{\Theta}_b, \bm{\Xi}, \bm{\Lambda}, \lambda, \bm{\Gamma})
        \]
        \State \quad \quad \textbf{Update regularization parameters:}\vspace{2mm}
        \State \quad \qquad  \textbf{where} \( s \gets +1 \) for ascent, \( s \gets -1 \) for descent
        \vspace{2mm}
        \State \quad \qquad \qquad \textbf{if} Uniform Sparsity \textbf{then}
        \[
        \lambda_{n+1} \gets \lambda_n + s \cdot \eta_{\lambda} \nabla_{\lambda} \mathcal{L}_b(\bm{\Theta}_b, \bm{\Xi}, \bm{\Lambda}, \lambda, \bm{\Gamma})
        \]
        \State \quad \qquad \qquad \textbf{else}
        \[
        \bm{\Gamma}_{n+1} \gets \bm{\Gamma}_n + s \cdot \eta_{\bm{\Gamma}} \nabla_{\bm{\Gamma}} \mathcal{L}_b(\bm{\Theta}_b, \bm{\Xi}, \bm{\Lambda}, \lambda, \bm{\Gamma})
        \]

        \State \quad \quad \textbf{Thresholding:} For each element $i$, set $\bm{\Xi}_{n+1}^i, \bm{\Lambda}_{n+1}^i \gets 0$ if $|\bm{\Xi}_{n+1}^i|, |\bm{\Lambda}_{n+1}^i| \leq \epsilon$
        \State \quad Increment iteration counter: $n \gets n + 1$
    \EndWhile
    
    \State \textbf{Output:}
    \State \quad Optimized parameters $\bm{\Xi}^*$ and $\bm{\Lambda}^*$
\end{algorithmic}
\end{algorithm}

\textit{Stochastic versus Full-Batch Optimization:} The ADAM-SINDy framework provides the flexibility to implement both a batching strategy through stochastic gradient descent and a full-batch optimization approach. In the stochastic case, optimization is performed on smaller, randomly sampled subsets of the overall dataset. This method optimizes computational resource utilization by partitioning the data into manageable chunks, thereby reducing memory requirements and accelerating the optimization process. The batch size must be sufficiently large to capture the dynamics of interest while remaining small enough to facilitate efficient computation. In contrast, full-batch optimization processes the entire dataset in a single iteration. This approach leverages all available information, potentially yielding more accurate and stable updates to the optimization parameters. Notably, ADAM optimization employs momentum, which can help avoid local minima even in the absence of stochasticity introduced by mini-batches.

While both methodologies share the same optimization strategy, a primary difference lies in the number of temporal data points considered in each optimization step. In the stochastic batching case, the batch size is typically smaller than the total number of available temporal data points. These data points are randomly sampled and shuffled, with the final batch potentially containing fewer points than the specified batch size. During implementation, temporal data is organized into rows based on the selected batch size, which dictates the number of rows in the ground truth data used in each optimization step. Then the candidate function matrix is constructed, reflecting the selected batch size through the number of rows it contains. Importantly, the dimensionality of the unknown nonlinear parameters and the coefficient matrix remains invariant, as the underlying dynamics of the system are independent of data partitioning based on time steps. Consequently, employing a batching strategy in the sparse regression problem enhances computational efficiency, reduces memory requirements, and facilitates effective manipulation of the matrix size.

\section{Joint Minimization vs.\ Min-Max Optimization}\label{appendix:comp}
In this section, we compare two optimization formulations considered within ADAM-SINDy: the \emph{min-max} formulation and the \emph{joint minimization} formulation. Both approaches aim to balance prediction accuracy and model sparsity but differ fundamentally in how sparsity is enforced. First, the min-max objective function is given by
\[
\min_{\mathbf{\Xi}, \mathbf{\Lambda}} \max_{\mathbf{\Gamma}} \left[ 
\left\| \dot{\mathbf{X}} - \mathbf{\Theta}(\mathbf{X}; \mathbf{\Lambda}) \mathbf{\Xi} \right\|_2^2 
+ \lambda \left\| |\mathbf{\Gamma}| \mathbf{\Xi} \right\|_1 
\right]\;,
\]
where the sparsity term is adversarially weighted by the variable $\mathbf{\Gamma}$. In contrast, the joint minimization formulation eliminates the adversarial structure by optimizing all variables cooperatively:
\[
\min_{\mathbf{\Xi}, \mathbf{\Lambda}, \mathbf{\Gamma}} \left[ 
\left\| \dot{\mathbf{X}} - \mathbf{\Theta}(\mathbf{X}; \mathbf{\Lambda}) \mathbf{\Xi} \right\|_2^2 
+ \lambda \left\| |\mathbf{\Gamma}| \mathbf{\Xi} \right\|_1 
\right]\;,
\]
treating $\mathbf{\Gamma}$ as a passive regularization weight rather than an adversarial one.
In both cases, the objective function consists of a data fidelity term (the \(\ell_2\) norm) and a sparsity-promoting term (the \(\ell_1\) norm). However, the role of $\mathbf{\Gamma}$ as either an adversary or a cooperative weight introduces fundamentally different optimization dynamics.

\subsection*{Theoretical analysis of the optimization frameworks}

To analyze the optimization dynamics, we consider the gradients of the objective function with respect to $\mathbf{\Xi}$ and $\mathbf{\Gamma}$:

\[\nabla_{\mathbf{\Xi}} L = -2 \mathbf{\Theta}^\top (\dot{\mathbf{X}} - \mathbf{\Theta} \mathbf{\Xi}) + \lambda \;|\mathbf{\Gamma}| \cdot\mathrm{sign}(\mathbf{\Xi}),\]
\[\nabla_{\mathbf{\Gamma}} L = \lambda \, |\mathbf{\Xi}| \cdot\mathrm{sign}(\mathbf{\Gamma}).\]
where, $\mathrm{sign}(z) = \{ +1 \text{ if } z>0, \; 0 \text{ if } z=0, \; -1 \text{ if } z<0 \}$. These expressions highlight the fundamental instability in the optimization dynamics when $\mathbf{\Gamma}$ is treated adversarially (e.g., min-max). The gradient $\nabla_{\mathbf{\Gamma}} L$ drives $\mathbf{\Gamma}$ to grow unbounded in directions that increase the sparsity penalty $\| |\mathbf{\Gamma}| \mathbf{\Xi}\|_1$, provided $\mathbf{\Xi}$ contains nonzero values. As $\mathbf{\Gamma}$ increases, the regularization term increasingly dominates $\nabla_{\mathbf{\Xi}} L$, subsequently driving $\mathbf{\Xi} \rightarrow 0$. This feedback loop leads to pathological dynamics in which $|\mathbf{\Gamma}| \rightarrow \infty$ and $\mathbf{\Xi} \rightarrow 0$, effectively suppressing model expressivity and hindering recovery of the correct terms. Importantly, this instability arises from the fact that the sparsity term is convex in both $\mathbf{\Xi}$ and $\mathbf{\Gamma}$. As a result, the overall objective lacks the convex-concave structure required for the existence of a well-defined saddle point. Specifically, a min-max problem admits a stable saddle point only when the objective is convex in the minimization variables and concave in the maximization variables. In our setting, the $\ell_1$ norm term is ``convex'' in both arguments $\mathbf{\Xi}$ and $\mathbf{\Gamma}$, rendering the adversarial maximization over $\mathbf{\Gamma}$ ill-posed. This contrasts with adversarial formulations successfully employed in other domains—such as self-adaptive PINNs~\cite{mcclenny2023self}—where adaptive weighting is applied to loss terms that are ideally driven to zero (e.g., initial or boundary condition errors). In sparse regression, however, the $\ell_1$ penalty should not be driven to zero through increasing $\Gamma$ unless $\mathbf{\Xi}$ is identically zero, making adversarial maximization of the penalty inherently self-defeating. Without a natural saturation point, $\mathbf{\Gamma}$ continues to grow, leading to complete sparsity and failure to recover meaningful terms.

We do acknowledge that there are methods for artificially introducing a saddle point into the min-max loss landscape that can benefit the optimization. For instance, hyperparameters such as weight decay, can prevent excessive growth of $\Gamma$ leading to a maximum sparsity enforcement. Alternative optimization algorithms, as opposed to ADAM, such as optimistic gradient descent~\cite{daskalakis2017training} or optimistic mirror descent~\cite{mertikopoulos2018optimistic}, introduce negative gradient components that can also yield saddle points for optimization. However, in the sparse regression setting, these additional parameters proved to be  sensitive and problem-specific through testing, limiting the robustness of the min-max formulation. Thus, even though the min-max optimization with these additional enhancements could solve these problems, we discourage its utilization.

In contrast, the joint minimization formulation avoids adversarial dynamics altogether, where the gradients $\nabla_{\mathbf{\Gamma}} L$ and $\nabla_{\mathbf{\Xi}} L$ act cooperatively to minimize the same objective. As a result, the gradient dynamics differ fundamentally. The parameter $\mathbf{\Gamma}$ no longer attempts to maximize the sparsity penalty but instead acts as a cooperative scaling factor on the $\ell_1$ regularization term. The minimization, in combination with the absolute value function on $\Gamma$, will always drive $\Gamma$ towards 0, differing from the unbounded growth in the min-max formulation. This movement towards zero is what allows for machine precision accuracy in nonlinear parameters as the impact of the $\ell_1$ penalty is reduced and as $\Gamma = 0$, we reach a least-squares problem. We can also verify that $\Gamma$ will stabilize at nonzero values for incorrect candidate functions from the above $\nabla_{\mathbf{\Gamma}} L$ equation and the thresholding step, preventing complete removal of the $\ell_1$ penalty. When an incorrect candidate function is zeroed out due to sequential thresholding, the gradient becomes exactly zero, preventing further updates to $\nabla_{\mathbf{\Gamma}} L$. Therefore, $\mathbf{\Gamma}$ saturates at a nonzero value for the incorrect terms. However, the algorithm allows the $\Gamma$ values associated with correct candidate terms to continue moving towards 0, further increasing the relative impact of the $\ell_2$ loss, while preventing the discovered model sparsity from decreasing. It is additionally noteworthy that the trajectory of $|\bm{\Gamma}|$ does not decrease monotonically, contrary to intuition from the analytical gradient $\nabla_{\mathbf{\Gamma}} L$. Instead, it exhibits oscillations over the training epochs as the true sign of $\mathbf{\Gamma}$ flips from positive to negative, since achieving an exact zero value of $\mathbf{\Gamma}$ is practically impossible without explicit clipping. This oscillatory behavior also further justifies the incorporation of a scheduler (dynamic learning rate) to decrease the learning rate at specific epochs, with the resulting oscillations further scaled by minimum step size of $\eta_\mathbf{\Gamma} \cdot \epsilon$. For $\mathbf{\Gamma}$ to consistently move closer to zero and thus increase the $\ell_2$ impact on remaining model candidates, one of these parameters ($\epsilon$ or $\eta_\mathbf{\Gamma}$) must decrease. Reducing $\epsilon$ risks allowing incorrect terms to re-enter the model; therefore, it is preferable to reduce the learning rate $\eta_\mathbf{\Gamma}$ instead.


Although this joint minimization approach may appear mathematically ambiguous, it is consistent with the design of an adaptive weighting strategy---penalizing only incorrect terms while sparing correct ones. The sparsity parameters converge to zero for the correct terms (facilitating machine precision), whereas they saturate at a nonzero value for the incorrect terms. While this approach avoids the ill-posedness associated with adversarial optimization, it may reduce the efficacy of sparsity enforcement, particularly when the regularization parameter $\lambda$ is small or the data fidelity term dominates the objective. In such cases, the learned weights $\mathbf{\Gamma}$ may fail to sufficiently penalize small-magnitude components in $\mathbf{\Xi}$, resulting in denser models. Nevertheless, the joint formulation offers a more stable and tractable optimization landscape, making it better suited for robust and interpretable sparse regression in practice.

\subsection*{Empirical analysis of the optimization frameworks}
\begin{figure}  
    \centering
    \includegraphics[width=.85\textwidth,height=\textheight,keepaspectratio]{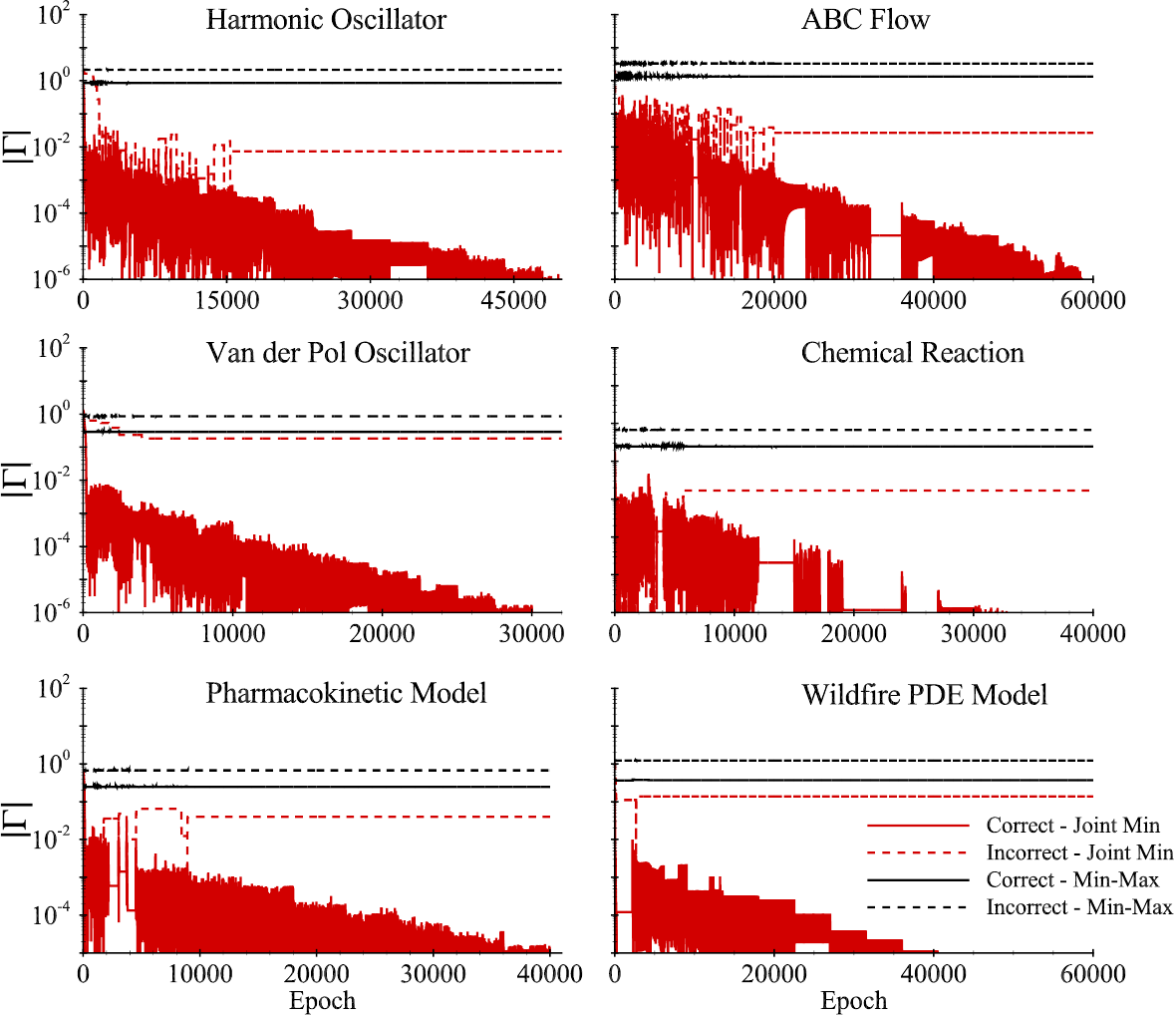}
    \caption{Evolution of $\mathbf{|\Gamma|}$ over training epochs for all numerical examples considered in this work. Each plot shows the $\mathbf{|\Gamma|}$ trajectory for both randomly selected correct and  incorrect terms, under both objective functions: joint minimization and min-max formulations.}

    \label{fig:gamma}
\end{figure}

To substantiate the theoretical discussion, we investigate the sparsity inducing $\bm{\Gamma}$ trajectory under both optimization formulations across all numerical examples considered in this work. Although the min-max formulation lacks a well-defined saddle point, we employ the \textit{optimistic} gradient ascent/descent method—an algorithmic strategy developed to improve stability for adversarial training~\cite{daskalakis2018limit}. This method introduces a negative momentum term based on past gradients into the current updates, effectively mitigating oscillations and reducing the risk of divergence. Specifically, optimistic ADAM~\cite{daskalakis2017training} is used for the maximization step, while vanilla ADAM is applied for the minimization step of the objective function.

A comparative analysis of the two objective formulations with respect to sparsity enforcement is presented in Fig.~\ref{fig:gamma}, where the evolution trajectories of $|\mathbf{\Gamma}|$ are plotted for two randomly selected candidate functions: one corresponding to a correct term and the other to an incorrect term. Across all experiments, the sparsity value of $\bm{\Gamma}$ associated with the incorrect term remains consistently higher than that of the correct term. This behavior aligns with the design of the adaptive weighting strategy, which aims to penalize incorrect candidates more strongly while applying minimal or no penalty to correct ones. Furthermore, despite identical initializations of $\bm{\Gamma}$, the min–max formulation imposes stronger sparsity constraints than the joint-minimization method, resulting in generally larger $\bm{\Gamma}$ values. Notably, this led to no candidate functions being selected under the min–max formulation, whereas the joint-minimization approach successfully identified correct terms as discussed in Sec.~\ref{sec3}. We further observe that replacing optimistic ADAM with vanilla ADAM for the maximization step intensifies sparsity enforcement; however, these results are omitted for brevity.

While the min–max formulation may appear advantageous in theory, its lack of convex-concave structure results in unstable adversarial dynamics that hinder practical sparse regression. The joint-minimization framework, by avoiding these adversarial effects, provides a more stable and reliable optimization landscape that still enforces sparsity effectively. Thus, it serves as the preferred strategy within the ADAM-SINDy framework throughout this work.

\bibliographystyle{unsrt}
\bibliography{sgd_reference}

\end{document}